\pdfoutput=1
\PassOptionsToPackage{table,xcdraw}{xcolor}
\documentclass[11pt]{article}

\usepackage{acl}

\usepackage{times}
\usepackage{latexsym}

\usepackage[T1]{fontenc}
\usepackage[utf8]{inputenc}

\usepackage{microtype}

\usepackage{inconsolata}
\usepackage{todonotes}
\usepackage{rotating}
\usepackage{graphicx}
\usepackage{balance} % for balancing columns on the final page
\usepackage{lscape}
\usepackage{amsmath}
\usepackage{amssymb}
\usepackage{amsthm}
\usepackage{float}
\usepackage{multirow}
\usepackage{algorithm}
\usepackage{algpseudocode}
\usepackage{tcolorbox}
\usepackage{booktabs}
\usepackage{pifont}
\usepackage{bbm}
\usepackage[normalem]{ulem}
\usepackage{hyperref}
\usepackage{algorithmicx}
\usepackage{dsfont}
\useunder{\uline}{\ul}{}
\newcommand{\KGCP}{\textsc{KGCP}}
\newcommand{\MCP}{\textsc{MCP}}
\newcommand{\ClusterCP}{\textsc{ClusterCP}}
\newcommand{\APS}{\textsc{APS}}
\newcommand{\RAPS}{\textsc{RAPS}}
\newcommand{\CondKGCP}{\textsc{CondKGCP}}

\newcommand{\Softmax}{SOFTMAX}

\newcommand{\CovGap}{CovGap}
\newcommand{\AvgSize}{AveSize}

\newcommand{\triple}[3]{\langle #1, #2, #3 \rangle}
\newcommand{\tr}[2]{\operatorname{tr}(#1, #2)}

\newcommand{\aTriple}{\mathit{tr}}

\newcommand{\Model}{M_{\theta}}
\newcommand{\RealSet}{\mathbb{R}}
\newcommand{\AnswerRank}{\operatorname{rank}_{\Model}(q, e)}

\newcommand{\partOf}[1]{\operatorname{part}(#1)}

\newcommand{\TSet}{\mathcal{T}}
\newcommand{\NamedTSet}[1]{\mathcal{T}_\mathrm{#1}}
\newcommand{\TrainingSet}{\NamedTSet{tr}}
\newcommand{\NegSet}{\NamedTSet{neg}}
\newcommand{\CalibrationSet}{\NamedTSet{cal}}
\newcommand{\TestSet}{\NamedTSet{test}}
\newcommand{\PartSet}{\TSet_{g}}

\newcommand{\Distribution}{\mathcal{P}}
\newcommand{\NonconformingScore}{\hat{s}_\epsilon}
\newcommand{\NonconformingScoreR}[1]{\NonconformingScore(\TSetR{#1})}
\newcommand{\PredictionSet}[1]{\hat{C}(#1)}
\newcommand{\PredictionSetMCP}[1]{\hat{C}_{\mathrm{MCP}}(#1)}
\newcommand{\PredicitonSetCondKGCP}{\hat{C}_{\mathrm{CondKGCP}}}

\newcommand{\Quantile}{\operatorname{quant}}

\newcommand{\Probability}[1]{\mathbb{P}(#1)}
\newcommand{\pred}[1]{\operatorname{pred}(#1)}
\newcommand{\TSetR}[1]{\CalibrationSet[#1]}

\newcommand{\similar}{\operatorname{sim}}

\newcommand{\FilteredE}[2]{E_q[#1 \leq #2]}
\newcommand{\vect}[1]{\boldsymbol{#1}}
\newcommand{\vectI}[1]{(\vect{#1})_i}

\DeclareMathOperator*{\argmax}{argmax}

\newcommand{\EnoughTriplesPredicates}{R_{\text{enough-data}}}
\newcommand{\InsuficientTriplesPredicates}{R_{\text{few-data}}}

\newcommand{\RankThreshold}{\hat{k}(g)}
\newcommand{\ScoreThreshold}{\hat{s}_{\epsilonG}(\PartSet)}
\newcommand{\epsilonG}{\epsilon'(g)}

\newcommand{\MiscoverageError}[1]{\epsilon_{g}^{#1}}

\newcommand{\CovR}{\mathrm{Cov}_r}

\algdef{SE}[DOWHILE]{Do}{doWhile}{\algorithmicdo}[1]{\algorithmicwhile\ #1}%

\newtheorem{theorem}{Theorem}

\newtheorem{proposition}{Proposition}

\newtheorem{corollary}{Corollary}
\algnewcommand{\LineComment}[1]{\State \(\triangleright\) #1}

\title{ Predicate-Conditional Conformalized Answer Sets for Knowledge Graph Embeddings }

\author{%
Yuqicheng Zhu\textsuperscript{1,2}, 
Daniel Hernández\textsuperscript{1},
Yuan He\textsuperscript{3},  
Zifeng Ding\textsuperscript{4},
Bo Xiong\textsuperscript{5},\\
\textbf{Evgeny Kharlamov\textsuperscript{2,6},} \textbf{Steffen Staab\textsuperscript{1,7}}
\\
\textsuperscript{1}University of Stuttgart, 
\textsuperscript{2}Bosch Center for AI, 
\\
\textsuperscript{3}University of Oxford, 
\textsuperscript{4}University of Cambridge, 
\textsuperscript{5}Stanford University, 
\\
\textsuperscript{6}University of Oslo, 
\textsuperscript{7}University of Southampton\\
\texttt{yuqicheng.zhu@de.bosch.com}\\
}

\begin{document}
\maketitle
\begin{abstract}
%Knowledge Graph Embedding (KGE) methods have shown effectiveness in various downstream tasks, yet uncertainty quantification for their prediction remains underexplored.
Uncertainty quantification in Knowledge Graph Embedding (KGE) methods is crucial for ensuring the reliability of downstream applications.
A recent work applies conformal prediction to KGE methods, providing uncertainty estimates by generating a set of answers that is guaranteed to include the true answer with a predefined confidence level.
However, existing methods provide probabilistic guarantees averaged over a reference set of queries and answers (\emph{marginal coverage guarantee}).
In high-stakes applications such as medical diagnosis, a stronger guarantee is often required: the predicted sets must provide consistent coverage per query (\emph{conditional coverage guarantee}).
%We observe that existing methods fail to meet such conditional coverage guarantee. 
We propose {\CondKGCP}, a novel method that approximates predicate-conditional coverage guarantees while maintaining compact prediction sets.
%balances the trade-off between conditional coverage probability and the size of the predicted sets. 
{\CondKGCP} merges predicates with similar vector representations and augments calibration with rank information. 
We prove the theoretical guarantees and demonstrate empirical effectiveness of {\CondKGCP} by comprehensive evaluations.

\end{abstract}

\section{Introduction}
% \begin{figure}[t!]
%     \centering
%     \includegraphics[width=0.9\linewidth]{images/cond_intro.pdf}
%     \caption{Coverage and frequency by predicate ID, illustrating the failure of {\KGCP} to provide adequate conditional coverage guarantees. The red line indicates the desired coverage level. The top plot shows that {\KGCP} provides lower coverage, especially for less frequent relations.}
%     \label{fig:intro}
% \end{figure}
Knowledge Graph Embeddings (KGE) encode entities and predicates as numerical vectors, enabling reasoning by exploiting similarities and analogies between entities and relations \citep{wang2017knowledge, biswas2023knowledge}. 
While KGE methods have demonstrated effectiveness in various downstream tasks such as link prediction
\cite{bordes2013translating, nickel2011rescal} and question answering \cite{saxena2020improving}, there remains uncertainty regarding the reliability of their predictions. 
Specifically, 
%since knowledge graphs (KGs) are inherently incomplete and sparse, 
KGE models fail to identify when the answers to a query are uncertain \cite{zhu2024predictive}.
%provide reliable predictions for certain queries  
%Identifying those uncertain queries is challenging and remains underexplored.
%for some queries, KGE models might not learn enough patterns to provide a reliable ranking. 
%And the plausibility scores returned by KGE methods do not have probabilistic interpretation \cite{Tabacof2020calibration}, therefore cannot express how confident the model is for the predictions. 

%\citet{zhu2024predictive} point out that substantially large proportion of queries have large uncertainty since the predicted rankings are very sensitive to small changes in training settings. 

%Specifically, given a query, the output of a KGE model is typically a ranking of all possible answers with a plausibility score assigned to each answer. Since the plausibility scores returned by KGE methods do not have probabilistic interpretation \cite{Tabacof2020calibration}, this kind of output cannot express how confident it is of a prediction.
\emph{Conformal prediction} is a framework to quantify uncertainty by providing a \emph{prediction set}—a set of possible solutions for a given task—that is guaranteed to cover the ground truth solution with a predefined confidence level \cite{vovk2005algorithmic}. 
By assigning a score to each possible solution, the method defines a threshold to choose the minimum number of elements for the prediction set to provide the coverage guarantee.
Thus, the size of the predicted set reflects the uncertainty of the predictions, with larger sets indicating higher uncertainty. 

Recently, \citet{zhu2024conformalized} introduced a method, Conformalized Knowledge Graph Embedding ({\KGCP}), which applies conformal prediction to quantify uncertainty in the predictions from KGE models. 
%for the link prediction task.
%Similar to estimating prediction intervals for regression problems \cite{dewolf2023valid}, 
%{\KGCP} provides a set of answers as output
%(can be seen as prediction interval for discrete values) 
%for a given query, within which the correct answer is guaranteed to fall with a specified probability (e.g. 95\%). 
%Intuitively, {\KGCP} tells us which entities to include in the prediction set to ensure, with high confidence (e.g. 95\%), that the true answer is covered by this set.
%The size of the predicted set reflects the uncertainty of the predictions, with larger sets indicating higher uncertainty.
%The effectiveness of the uncertainty quantification methods is therefore measured by (1) coverage: whether the prediction set cover the true answer with the specified probability (2) average size: whether the method provides reasonably small prediction sets.
They show that {\KGCP} provides \emph{marginal coverage guarantees}, ensuring that the prediction sets meet the desired confidence level on average across all queries.
%cover the true answer with desired probability on average across all possible test queries. 
%However, KGs often exhibit significant heterogeneity in the semantics and complexity of predicates. 
%For instance, some predicates represent transitive relationships (e.g., "isAncestorOf"), while others are purely categorical (e.g., "hasColor"). 
However, predictive uncertainty may vary substantially across predicates, necessitating tailored coverage guarantees conditioned on predicates (\emph{predicate-conditional coverage guarantees}). 
Such conditional guarantees are especially crucial for real-world applications where specific subgroups demand reliable uncertainty estimates.
%coverage guarantees are needed for specific subgroups. 
For instance, in a medical diagnosis system leveraging KGE, predicates like ``contraindicated\_for'' (indicating that a treatment is not recommended for certain patients) and ``has\_symptom'' (indicating that a specific disease or condition is associated with certain symptoms) may require different thresholds to achieve prediction sets with the desired confidence level.
A shared threshold might fail to cover the true answer for the predicate ``contraindicated\_for'', as it is often associated with fewer triples and demands a higher threshold. 

%to higher uncertainty in predictions.
%yet these predicates often appear in only a few triples therefore might result in more uncertain predictions returned by KGE models. 
%An uncertainty quantification method that does not ensure conditional coverage could perform well overall but fail to reliably cover true answers for these rare but crucial predicates.
%However, we observe that 
%{\KGCP} fail to have coverage guarantees for specific subgroup of queries with different level of uncertainty.
%, as shown in Figure \ref{fig:intro}.
%for instance, queries with less frequent predicates (As shown in Figure \ref{fig:intro}).

Conditional coverage guarantee can be achieved by performing conformal prediction at the subgroup level \cite{vovk2005algorithmic}. 
However, the highly imbalanced distribution of triples across predicates in KGs \cite{xiong2018oneshot} poses challenges, often resulting in prediction sets that are either overly large or fail to cover the true answer \cite{ding2024class, shi2024conformal}.
To address this limitation, we propose {\CondKGCP}, a method designed to approximate predicate-conditional guanratee while maintaining compact prediction sets.
The key components of {\CondKGCP} are as follows:
(1) it \textbf{merges predicates} with similar vector representation to increase the number of calibration triples available for reliable subgroup-level conformal prediction, and (2) it introduces a \textbf{dual calibration schema} that combines score calibration with rank calibration to exclude noisy answer entities, thereby reducing the size of prediction sets.

We provide theoretical guarantees that {\CondKGCP} achieves conditional coverage probabilities tightly centered around the desired confidence level and that the dual calibration schema reduces expected prediction set sizes under certain conditions. Empirically, we demonstrate that {\CondKGCP} outperforms five baseline methods, achieving a superior trade-off between conditional coverage probability and prediction set size across commonly used benchmark datasets. 

\section{Related Work}
The majority of KGE methods aim to improve model performance by capturing relational patterns through more expressive embedding spaces, such as complex \cite{trouillon2016complex, sun2019rotate}, hyperbolic \cite{xiong2022ultrahyperbolic}, or probabilistic spaces \cite{he2015learning}. By enabling richer representations, these methods have shown strong performance across downstream tasks including query answering \cite{ren2020query2box, he2024generating, he2025dage}, recommendation \cite{sun2018recurrent}, and image classification \cite{zhou2024visual}. Despite these successes, uncertainty quantification in KGE remains largely underexplored.
Most uncertainty quantification methods for KGE calibrate the plausibility scores generated by the models \cite{Tabacof2020calibration, Safavi2020calibration}. However, these methods lack formal guarantees for the resulting probabilities. In contrast, \citet{zhu2024conformalized} introduce an approach that 
provides formal statistical guarantees.

It is well-established that no prediction interval can achieve a conditional coverage guarantee in a finite sample without additional assumptions about the data distribution \cite{vovk2012conditional, foygel2021limits}. Consequently, many of the existing works provide coverage guarantees conditioned on specific subgroups, such as class-conditional coverage guarantees \cite{ding2024class, shi2024conformal}.
Two main strategies have been proposed to improve conditional coverage probabilities. The first involves modifying the nonconformity measure. For instance, \citet{romano2020classification} enhance conditional coverage by defining cumulative probability of ground truth as nonconformity score, though their approach often results in larger prediction sets. 
To reduce the size of prediction sets, \citet{Angelopoulos2021image} introduce a regularization term in the nonconformity score.
The second strategy leverages subgroup-level conformal prediction. \citet{vovk2005algorithmic} propose Mondrian Conformal Prediction, which performs conformal prediction within specific subgroups. Building on this, \citet{ding2024class} cluster calibration points based on the distribution of nonconformity scores, balancing the trade-off between conditional coverage probability and the size of the prediction sets. Additionally, \citet{shi2024conformal} further optimize prediction set sizes by incorporating rank information during the calibration step. However, these methods focus on classification, while our approach targets KGE-based link prediction, which is more challenging due to the large number of potential answers and the highly imbalanced triple distribution across predicates.

\section{Preliminaries}

\subsection{Notations}

Given two finite sets $E$ and $R$ whose elements are called \emph{entities} and \emph{predicates}, a \emph{knowledge graph (KG)} is a subset of $E \times R \times E$, whose elements are known as \emph{triples}.
A \emph{query} $q$ is either an expression of the form $\triple{h}{r}{?}$ or $\triple{?}{r}{t}$, where $h,t \in E$, $r \in R$, and the question mark denotes the missing entity that we need to find.
Given a query $q$, $\pred{q}$ is the predicate of the query, and $\tr{q}{e}$ is the triple that results from assuming that $e$ is an answer to the query. That is, $\pred{\triple{?}{r}{t}} = r$, $\pred{\triple{h}{r}{?}} = r$, $\tr{\triple{?}{r}{t}}{e} = \triple{e}{r}{t}$ and $\tr{\triple{h}{r}{?}}{e} = \triple{h}{r}{e}$.

% I The following is an alternative notation. The goal of this notation is to avoid enumerating the elements in the training set, because the order does not matter. I am still working on this paragraph. I will continue after a walk.
% BEGIN alternative

A \emph{query-answer} set $\TSet$ is a finite set of pairs $(q,e)$ where $q$ is a query, $e \in E$ is an answer to the query. 
%and given two elements $(q_1, e_1)$ and $(q_2, e_2)$ of set $\TSet$, if $\tr{q_1}{e_1} = \tr{q_2}{e_2}$ then $q_1 = q_2$ and $e_1 = e_2$. 
Abusing notation, given a triple $\aTriple$, we write $\aTriple \in \TSet$ if there is a pair $(q, e) \in \TSet$ such that $\tr{q}{e} = \aTriple$. 
%and we say that two query-answer sets $\TSet_1$ and $\TSet_2$ are \emph{disjoint} if there is no triple $\aTriple$ such that $\aTriple \in \TSet_1$ and $\aTriple \in \TSet_2$. 
We use the names $\TrainingSet$, $\NegSet$, $\CalibrationSet$, and $\TestSet$ for the query-answer sets that are usually called \emph{training set}, \emph{negative triples set}, \emph{calibration set}, and \emph{test set}. 

KGE methods train KGE models $\Model : E \times R \times E \to \RealSet$ with parameters $\theta$ using a given training set $\TrainingSet$ sampled from a distribution $\Distribution$, whose elements are called \emph{positive triples}, and a set $\NegSet$ of negative triples, disjoint with $\TrainingSet$. The learned model assigns scores to triples indicating their plausibility. It gives higher scores to the positive triples, and lower scores to the negative triples \cite{bordes2013translating, nickel2011rescal}.

The performance of a KGE model is typically evaluated by the rank of answers in the test set. Given a pair $(q, e) \in \TestSet$, the \emph{rank} of answer $e$ to query $q$ predicted by $\Model$, denoted $\AnswerRank$ is the size of the set of elements $E\ni e'$ such that $\Model(\tr{q}{e'}) \geq \Model(\tr{q}{e})$.
Smaller rank values indicate a better model performance.

\subsection{Conformalized KGE}

Given a KGE model $\Model$ trained on $\TrainingSet$, a pair $(q, e) \in \TestSet$, and a user-specified error rate $\epsilon \in [0,1]$, {\KGCP} \cite{zhu2024conformalized} provides a set of entities, 
%$\hat C(q)\subseteq E$, 
which is guaranteed to contain $e$ with a probability of at least $1-\epsilon$.
In this section, we provide background on the method.

% Suppose a KGE model is trained on $\mathcal{T}_{tr}$. Given a test query $q_{test}$ and a user-specified error rate $\epsilon$, {\KGCP} \cite{zhu2024conformalized} provides a set of entities $\hat C(q_{test})\subseteq E$, which is guaranteed to contain $e_{test}$ with a probability of at least $1-\epsilon$.

A \emph{nonconformity measure} $S: E \times R \times E \rightarrow \mathbb{R}$ quantifies how unusual a triple is with respect to the training set. This measure is typically derived from a pre-trained KGE model, e.g., $S(tr) = -M_\theta(tr)$ \cite{zhu2024conformalized}. 
Based on the nonconformity measure, the procedure of conformal prediction consists of two steps: 
%i.e., the negative plausibility score of a triple 

\textbf{Calibration Step}: given a number $\tau \in [0,1]$ and a finite set $A \subseteq \mathbb{R}$, the $\tau$-quantile of $A$, 
 denoted $\Quantile(\tau, A)$, is infimum of the set of elements $a \in A$ such that
$|\{b\in A:b\leq a\}| / |A| \geq \tau$.
%is the smallest $a \in A$ such that a $\tau$-fraction of the elements in $A$ are less than or equal to $a$. 
Given a query-answer set $\TSet$, the empirical \emph{quantile of nonconformity scores} is:
\begin{align}
  %\TauE{\TSet} &= \frac{\lceil(|\TSet|+1)(1-\epsilon)\rceil}{|\TSet|} ,\\
  \label{eq:threshold}
  \NonconformingScore(\TSet) &= \Quantile\left(\frac{\lceil(|\TSet|+1)(1-\epsilon)\rceil}{|\TSet|}, \TSet\right).
\end{align}
Given the calibration set $\CalibrationSet$, for a target coverage $1-\epsilon$, we obtain the corresponding empirical quantile of nonconformity scores, $\NonconformingScore(\CalibrationSet)$. 
% Let $\mathcal{T}_{cal}=\{tr_i\}_{i=1}^m$ denote a calibration set. 
% For a target coverage $1-\epsilon$, we obtain the corresponding empirical quantile of nonconformity scores on $\mathcal{T}_{cal}$: 
% \begin{align}
%   \tau_\epsilon &= \frac{\lceil(|\CalibrationSet|+1)(1-\epsilon)\rceil}{|\CalibrationSet|} ,\\
%   \label{eq:threshold}
%   \NonconformingScore &= \Quantile\left(\tau_\epsilon, \CalibrationSet\right).
% \end{align}
% \begin{equation}\label{eq:threshold}
%      \hat s_{\epsilon} = \Quantile(\frac{\lceil(m+1)(1-\epsilon)\rceil}{m}, \{S(tr_i)\}_{i=1}^m)
% \end{equation}

\textbf{Set Construction Step}: 
Given a threshold $s$, and a query $q$, we define the set $\FilteredE{S}{s}$ as follows:
\begin{align}
    \FilteredE{S}{s} = \{ e \in E : S(\tr{q}{e}) \leq s \}.
\end{align}
The \emph{prediction set} for a test query $q$, denoted $\PredictionSet{q}$, is then constructed by including all answer entities that have nonconformity scores smaller than the threshold $\NonconformingScore(\CalibrationSet)$:
\begin{equation}\label{eq:predset}
    \PredictionSet{q} = \FilteredE{S}{\NonconformingScore(\CalibrationSet)}.
\end{equation}

% \textbf{Set Construction Step}: The prediction set for a test query $q_{test}$ is then constructed by including all answer entities that have nonconformity scores smaller than the threshold $\hat s_\epsilon$:
% \begin{equation}\label{eq:predset}
%     \hat C(q_{test}) = \{e\in E:S(tr(q_{test},e))\leq\hat s_{\epsilon}\}.
% \end{equation}

\begin{theorem}[\citet{zhu2024conformalized}]\label{cor:guarantee}
Suppose the triples in $\TrainingSet$, $\CalibrationSet$ and $\TestSet$ are drawn independent and identically distributed (i.i.d) from the underlying distribution $\mathcal{P}$. For every element $(q, e) \in \TestSet$, the probability of $e$ to being included in the prediction set of $q$ satisfies the following bounds:
(i) $\Probability{e \in \PredictionSet{q}}\geq 1-\epsilon$, and
(ii) if there is no tie in the set of scores of the triples in $\CalibrationSet$, then $\Probability{e \in \PredictionSet{q}} \leq
  1-\epsilon+\frac{1}{|\CalibrationSet| + 1}$.
% \begin{align}
%   \label{eq:KGCP_guarantee}
%   1-\epsilon \leq
%   \Probability{e \in \PredictionSet{q}}.
% \end{align}
% Furthermore, if there is no ties in $S(\CalibrationSet)$, then we also have the following upper bound:
% \begin{align}
%   \label{eq:KGCP_guarantee}
%   1-\epsilon \leq
%   \Probability{e \in \PredictionSet{q}} \leq
%   1-\epsilon+\frac{1}{|\CalibrationSet| + 1} .
% \end{align}
\end{theorem}

% \begin{corollary}[\citet{zhu2024conformalized}]\label{cor:guarantee}
% Suppose the triples in $\mathcal{T}_{tr}$, $\mathcal{T}_{cal}$ and $tr(q_{test}, e_{test})$ are drawn independent and identically distributed (i.i.d) from the underlying distribution $\mathcal{P}$, the prediction set $\hat C(q_{test})$ is guaranteed to cover the true answer with at least probability $1-\epsilon$ 
% \begin{equation}\label{eq:KGCP_guarantee_lower}
%     \mathbb{P}(e_{test}\in \hat C(q_{test}))\geq 1-\epsilon.
% \end{equation}
% Furthermore, if there is no ties in $\{S(tr_i)\}_{i=1}^m$, then we also have the following upper bound:
% \begin{equation}\label{eq:KGCP_guarantee_upper}
%     \mathbb{P}(e_{test}\in \hat C(q_{test}))\leq 1-\epsilon+\frac{1}{m+1}.
% \end{equation}
% \end{corollary}

\section{Conditional Conformal Prediction for Knowledge Graph Embedding (\CondKGCP)}
The goal of this paper is to approximate predicate-conditional coverage guarantee.
Given a pair $(q, e) \in \TestSet$, and an arbitrary predicate $r \in R$, Equation~\eqref{eq:pre_guarantee} defines the \emph{predicate-conditional coverage guarantee}, which ensures that the true answer $e$ is included in the prediction set of query $q$ with a probability of at least $1-\epsilon$.
%improve {\KGCP} in terms of predicate-conditional coverage probability. 
% For a test query $q_{test}$ and its corresponding answer $e_{test}$, we say that the prediction set $\hat C(q_{test})$ satisfies \emph{conditional coverage guarantee} at level $1-\epsilon$ if
% \begin{equation}\label{eq:cond_guarantee}
%     \small\mathbb{P}(e_{test}\in \hat C(q_{test})|q_{test})\approx 1-\epsilon.
% \end{equation}
% However, it is known that there is no finite-length prediction interval can achieve conditional coverage as defined by Equation \ref{eq:cond_guarantee}, without invoking further assumptions on $\mathcal{P}$ \cite{vovk2012conditional, lei2014distribution}. Therefore, in this paper, we focus on a more pragmatic yet relevant special case: predicate-conditional coverage of conformalized KGE methods. 
% Let $Pre(q_{test})\in R$ denote the predicate of the test query. The predicate-conditional coverage guarantee ensures that the true answer is included in the prediction set with a probability of at least $1-\epsilon$ for each predicate $r \in R$, as defined by:
%
\begin{equation}\label{eq:pre_guarantee}
    \Probability{e \in \PredictionSet{q} \mid \pred{q} = r}\geq 1-\epsilon
\end{equation}
%
% \begin{equation}\label{eq:pre_guarantee}
%     \mathbb{P}(e_{test}\in \hat C(q_{test})|Pre(q_{test})=r)\geq 1-\epsilon, \forall r\in R
% \end{equation}
%
Equation~\eqref{eq:pre_guarantee} does not necessarily hold for {\KGCP} because the predictive uncertainty and the nonconformity score distribution can vary dramatically across predicates, which violates the i.i.d assumption in Theorem~\ref{cor:guarantee}. 

To have the guarantee in Equation \eqref{eq:pre_guarantee}, a method called Mondrian Conformal Prediction ({\MCP}) \cite{vovk2005algorithmic} performs conformal prediction separately for each predicate.
Given a subset $A \subseteq R$, let $\TSetR{A}$ be the query-answer subset of $\CalibrationSet$ such that $\aTriple \in \TSetR{A}$ if and only if $\pred{\aTriple} \in A$. The prediction set for the method {\MCP} is defined as:
\begin{align}
  \label{eq:mcp_set}
  \PredictionSetMCP{q} &= \FilteredE{S}{\NonconformingScoreR{\{r\}}}.
\end{align}
%
% To have the guarantee in Equation \eqref{eq:pre_guarantee}, a method called Mondrian Conformal Prediction ({\MCP}) \cite{vovk2005algorithmic} performs conformal prediction separately for each predicate.
% Let $\mathcal{I}^r=\{i\in[m]: Pre(q_i)=r\}$ represent the indices of queries in the calibration set that correspond to predicate $r$. The answer set for the method {\MCP} is defined as:
% %
% \begin{equation}\label{eq:mcp_set}
%     \hat C_{MCP}(q_{test})= \{e:S(tr(q_{test},e))\leq\hat s^r_\epsilon\},
% \end{equation}
% where
% \begin{equation}\label{eq:mcp_tau}
%      \hat s^r_\epsilon = \Quantile(\frac{\lceil(|\mathcal{I}^r|+1)(1-\epsilon)\rceil}{|\mathcal{I}^r|}, \{S(tr_i)\}_{i\in\mathcal{I}^r}).
% \end{equation}
%
However, it is well known that most predicates in KGs are associated with very few triples \cite{xiong2018oneshot}, resulting in small sets $\TSetR{\{r\}}$. This leads to unstable thresholds $\NonconformingScore(\TSetR{\{r\}})$, which in turn causes prediction sets to become overly large or fail to cover the ground truth.

To address these issues, we propose: (1) merging predicates to increase the number of triples in the calibration set, and (2) augmenting the calibration process with rank information to reduce the size of prediction sets.

\subsection{Predicate Merging}\label{sec:merge}
To obtain a threshold $\NonconformingScoreR{\{r\}}$ that reliably covers the true answer with desired probability, it is necessary to have a sufficiently large set $\TSetR{\{r\}}$ \cite{vovk2005algorithmic, ding2024class}. To increase $\TSetR{\{r\}}$, we merge predicates with highly similar vector representations. 
The rationale is that such predicates are likely to have similar distributions of nonconformity scores and, consequently, similar $\NonconformingScoreR{\{r\}}$. 
Formally, we aim to partition the set $R$ such that each part corresponds to a subset of $\CalibrationSet$ whose triples share similar predicates and is large enough to determine a reliable threshold for constructing prediction sets.
To define a set partition, we propose Algorithm~\ref{alg:set-partition}, 
which first places all predicates with enough data into separate partitions and then assigns predicates with few data to the partition of the most similar predicate.

% (Daniel) I will contine after lunch.

% \begin{algorithm}
%   \caption{Predicate Set Partition}\label{alg:set-partition}
%   \begin{algorithmic}
%   \Require
%     The set $R$ of predicates, a natural number $0 \leq \phi \leq |R|$ such that there is a predicate $r \in R$ with $|\TSetR{\{r\}}| > \phi$, and a similarity function $\similar$ for pairs of predicates.
%   \Ensure A set partion $P$ of $R$ such that every part $A \in R$ satisfies $|\TSetR{A}| \geq \phi$.
%   \State $R_1 \gets \{ r \mid r \in R \text{ and } |\TSetR{\{r\}}| \geq \phi \}$
%   \State $R_2 \gets \{ r \mid r \in R \text{ and } |\TSetR{\{r\}}| < \phi \}$
%   \State $A_r \gets \{ r \}$, for each $r \in R_1$
%   \For{$r \in R_1$}
%     \State $r'' \gets \argmax_{r' \in R_1} \similar(r, r')$
%     \State $A_{r''} \gets A_{r''} \cup \{r\}$
%   \EndFor
%   \State $P \gets \{ A_r \mid r \in R_1 \}$
%   \end{algorithmic}
% \end{algorithm}

\begin{algorithm}
  \caption{Predicate Set Partition}\label{alg:set-partition}
  \begin{algorithmic}
  \Require
    The set of predicates $R$, a natural number $\phi \leq \max_{r \in R}|\TSetR{\{r\}}|$, and a similarity function $\similar$ for pairs of predicates.
  \Ensure A set partition $P$ of $R$ such that every part $A \in P$ satisfies $|\TSetR{A}| \geq \phi$.
  \State $\EnoughTriplesPredicates \gets \{ r \mid r \in R \text{ and } |\TSetR{\{r\}}| \geq \phi \}$
  \State $\InsuficientTriplesPredicates \gets \{ r \mid r \in R \text{ and } |\TSetR{\{r\}}| < \phi \}$
  \State $\partOf{r} \gets \{ r \}$, for each $r \in \EnoughTriplesPredicates$
  \For{$r' \in \InsuficientTriplesPredicates$}
    \State $r \gets \argmax_{r'' \in \EnoughTriplesPredicates} \similar(r', r'')$
    \State $\partOf{r} \gets \partOf{r} \cup \{r'\}$
  \EndFor
  \State $P \gets \{ \partOf{r} \mid r \in \EnoughTriplesPredicates \}$
  \end{algorithmic}
\end{algorithm}

% Formally, we apply a merging algorithm $f: R \rightarrow \{1, \dots, G\}$ that assigns each predicate $r\in R$ to one of $G$ merged groups based on vector representations learned by a KGE model.
% Let $v(\cdot)$ be the embedding function that maps predicates to their vector representations, 
% and let $sim(\cdot, \cdot)$ denote a similarity function. 
% Given a size threshold $\phi\in\mathbb{N}$, for each predicate $r\in R$ with $|\mathcal{I}^r|<\phi$, 
% we identify the most similar predicate $r^*$ that satisfies $|\mathcal{I}^{r^*}|\geq\phi$:
% \begin{equation}
%     r^* = \arg\max_{r'\in R: |\mathcal{I}^{r'}|\geq\phi}sim(v(r), v(r')).
% \end{equation}
% The predicate $r$ is then merged into the same group as $r^*$.

\noindent
Given a part $g$ of the partition $P$ defined by Algorithm~\ref{alg:set-partition}, let $\PartSet$ be the subset of $\CalibrationSet$ consisting of all triples whose predicates belong to $g$.

In this work, we use negative Manhattan distance as the similarity measure. Let $\vectI{x}$ denote the $i$-th dimension of the vector $\vectI{x}$ representation of a predicate $x$, and $d$ denote the number of dimensions. The similarity function is defined as
\begin{equation}
    \similar(a,b) = -\sum_{i=1}^d|\vectI{a} - \vectI{b}|.
\end{equation}

\subsection{Dual Calibration Schema}\label{sec:dual_cal}

Prediction sets tend to be larger when conformal prediction is performed at the subgroup level due to the reduced number of calibration triples available for each subgroup compared to {\KGCP}.
To address this, and drawing inspiration from the recent work of \citet{shi2024conformal}, we reduce the size of prediction sets by constructing prediction sets using a dual calibration schema that combines score calibration and rank calibration.

% Let $\mathcal{I}^g=\{i\in [g]: f(r)=g \}$ denote the indices of queries in the $g$-th merged predicate group.
Given a query $q$ with predicate belonging to $g$, the prediction set generated by {\CondKGCP} is:
\begin{align}
    &\PredicitonSetCondKGCP(q)=\\
    &\notag\Big\{ e \in \FilteredE{S}{\ScoreThreshold}: \\
    &\notag\qquad\qquad\qquad\AnswerRank \leq \RankThreshold \Big\}.
\end{align}
It depends on two parameters, namely, the \emph{score threshold} $\ScoreThreshold$ and the \emph{rank threshold} $\RankThreshold$, which we will define in the remainder of this section.
%The prediction sets are defined as follows:
% \begin{align}
%     &\hat C_{CondKGCP}(q_{test})\\
%     &=\{e:S(tr(q_{test},e))\leq\hat s^g_{\epsilon'}, r_{M_\theta}(q_{test}, e)\leq\hat k(g)\}\notag,
% \end{align}
% where $s^g_{\epsilon'}$ is the score threshold and $\hat k(g)$ is the rank threshold.

\textbf{Rank Calibration.} Recall that $\AnswerRank$ is the rank of answer $e$ given query $q$. We define the \emph{miscoverage error of top-$k$ prediction set} for the part $g \in P$, denote $\MiscoverageError{k}$, as follows:
\begin{equation}
    \epsilon_{g}^k = \Probability{\AnswerRank > k \mid \pred{q} \in g}
\end{equation}
The rank threshold $\hat k(g)$ is selected such that $\MiscoverageError{\hat{k}(g)} < \epsilon$ to satisfy the coverage guarantee. However, achieving a smaller $\epsilon_g^{\hat{k}(g)}$ requires a larger $\hat k(g)$, which leads to larger prediction sets.
To minimize the size of the prediction sets, we choose 
%the smallest $k$: % such that the top-$k < \epsilon$:
\begin{equation}
    \RankThreshold = \min{\{k:\epsilon_{g}^k<\epsilon\}}.
\end{equation}

\textbf{Score Calibration.} We further apply a score threshold $\ScoreThreshold$ for the entities that are ranked within top-$\RankThreshold$, where $\epsilonG = \epsilon - \gamma\epsilon_{g}^{\hat k(g)}$ and $\gamma$ is a hyperparameter.

Intuitively, the rank threshold $\RankThreshold$ filters out answer entities with large rank positions (high $\AnswerRank$), ensuring that {\CondKGCP} performs score thresholding only on a subset of reliable test triples \cite{shi2024conformal}. The hyperparameter $\gamma$ balances the trade-off between the conditional coverage guarantee and the size of prediction sets (see a detailed explanation in the next section).

\section{Coverage \& Size Guarantees}
In this section, we will show the conditional coverage guarantee and size reduction guarantee of {\CondKGCP}. All proofs are in Appendix \ref{app:proof}.

\begin{proposition}[Conditional Coverage Guarantee]\label{prop:cov}
Let $q$ be a query, $e$ be its answer entity and $p$ be the conditional coverage probability of {\CondKGCP} 
$p=\mathbb{P}(e\in \PredicitonSetCondKGCP(q)\mid \pred{q} \in g)$.
Given a user-specified error rate $\epsilon$ and a $\gamma\in[0,1]$, we have the following bounds for all parts $g \in P$:
\begin{align}
    p \geq 1-\epsilon-(1-\gamma)\epsilon_g^{\hat k(g)},
\end{align}
and if there is no tie in the set of nonconformity scores of the triples in $\PartSet$, then
\begin{align}
    p \leq 1-\epsilon+\gamma\epsilon_g^{\hat k(g)}+\frac{1}{|\PartSet|+1}.
\end{align}
\end{proposition}
This proposition shows within each part $g$, the conditional coverage probability is close to $1-\epsilon$ with small controlled deviations. The deviation is governed by two "slack" terms: (1) the miscoverage error of rank calibration $\epsilon_g^{\hat k(g)}$ and (2) a finite-sample correction term $\frac{1}{|\PartSet|+1}$ to handle ties. Both terms are very small, $\epsilon_g^{\hat k(g)}$ is guaranteed to be smaller than $\epsilon$ by the way we select $\hat k(g)$ in Section \ref{sec:dual_cal}; $\frac{1}{|\PartSet|+1}$ is also guaranteed to be smaller than $\frac{1}{\phi+1}$ since we make sure that every part has at least $\phi$ triples in Algorithm \ref{alg:set-partition}.

Note that $\gamma$ does not affect the width of the coverage bounds but controls their asymmetry:
a larger $\gamma$ allows more deviation on the lower bound, while a smaller $\gamma$ does so for the upper bound. 
Furthermore, $\gamma$ influences the construction of the prediction sets by adjusting the score threshold via $\epsilon'(g)$:
A larger $\gamma$ reduces $\epsilon'(g)$, raising the threshold and yielding larger prediction sets, whereas a smaller $\gamma$ results in smaller prediction sets.
Thus, $\gamma$ essentially deals with the trade-off between conditional coverage probability and the size of prediction sets. 

%Proposition \ref{prop:cov} show that the lower bound of the conditional coverage probability provided by {\CondKGCP} is guarantee to be very close to the desired coverage probability $1-\epsilon$, since $\epsilon_g^{\hat k(g)}<\epsilon$ and $1-\gamma\in [0,1]$, the upper bound depend on the size of calibration set for the particular subgroup. Note by merging procedure described in Section \ref{sec:merge}, we obtain tighter upper bound compared to $MCP$.

% We name the method of {\CondKGCP} without rank calibration as Subgroup Conformal Prediction $SCP$, i.e. with prediction set
% \begin{equation}
%     \hat C_{SCP}(q_{test})= \{e:S(tr(q_{test},e))\leq\hat s^g_\epsilon\},
% \end{equation}
% where
% \begin{equation}\label{eq:answerset}
%      \small\hat s_{\epsilon}^g = \Quantile(\frac{\lceil(|\mathcal{I}^g|+1)(1-\epsilon)\rceil}{|\mathcal{I}^g|}, \{S(tr_i)\}_{i\in\mathcal{I}^g}).
% \end{equation}
% We can prove that the expected size of prediction sets provided by {\CondKGCP} is guaranteed to be smaller than $SCP$ under certain condition based on \cite[Lemma 4.2]{shi2024conformal}
\begin{corollary}[\citet{shi2024conformal}]
    Suppose $\epsilon'(g)$ and $\hat k(g)$ satisfy both following conditions
    \begin{equation}
        \hat k(g)\in\Big\{k:\epsilon_g^{\hat k(g)}<\epsilon\Big\}; 0\leq\epsilon'(g)\leq\epsilon-\epsilon_g^{\hat k(g)},
    \end{equation}
    the rank calibration guarantee to shrink the prediction sets, if for a query $q$ and any $e'\in E$:
    \begin{align}\label{eq:cond}
        \mathbb{P}_{q}\Big(S&(tr(q, e'))\leq\ScoreThreshold, \\\notag
        &\operatorname{rank}_{M_\theta}(q, e')\leq\hat k(g)\Big)\\\notag
        \leq\mathbb{P}&_{q}\Big(S(tr(q, e'))\leq\hat s_{\epsilon}(\PartSet)\Big)
    \end{align}
\end{corollary}
Intuitively, the dual calibration schema tends to include less answer entities with high rank from KGE models thus reduce the size of prediction sets \cite{shi2024conformal}. The corollary demonstrate it is true in theory under the condition of Equation (\ref{eq:cond}). We empirically verify the condition on benchmark datasets, the results in Appendix \ref{app:verify} show the practical utility of this corollary.

% However, when $|\mathcal{I}^r|$ is small, the resulting threshold $\hat s^r$ can be unstable, leading to overly large or erratic prediction sets \cite{ding2024class}. Our method, denoted as {\CondKGCP}, addresses this issue by grouping predicates with similar nonconformity score distributions. This strikes a balance between the granularity of $PreSpec$ and the desired coverage property of {\KGCP}.

% Specifically, we summarize the empirical distribution of nonconformity scores for each predicate as a vector of quantiles, forming an embedding vector $z^r \in \mathcal{Z}$. For a predicate $r$, the embedding vector is defined as:
% \todo[inline]{use density estimation}
% \begin{align*}
%     z^r=[&\Quantile(\tau, \{S(tr_i)\}_{i\in\mathcal{I}^r}):\\
%     &\tau\in\{0.5, 0.6, 0.7, 0.8, 0.9, 1.0\}]
% \end{align*}

% We then apply a clustering algorithm $f: \mathcal{Z} \rightarrow {1, \dots, M}$ to group the predicates based on their quantile vectors, where $M$ is the number of clusters.

% Let $\mathcal{I}^m=\{i\in [n]: f(z^r)=m \}$ denote the indices of queries in the $m$-th cluster. The prediction set for the {\CondKGCP} method is constructed as:
% \todo[inline]{non-parametric clustering}
% \begin{equation}
%     \hat C_{Cluster}(q_{n+1})= \{e:S(tr(q_{n+1},e))\leq\hat s^m\},
% \end{equation}
% where
% \begin{equation}\label{eq:answerset}
%      \small\hat s^m = \Quantile(\frac{\lceil(|\mathcal{I}^m|+1)(1-\epsilon)\rceil}{|\mathcal{I}^m|}, \{S(tr_i)\}_{i\in\mathcal{I}^m}).
% \end{equation}

\begin{table*}[h!]
\centering
\resizebox{\textwidth}{!}{%
\begin{tabular}{ccccc||ccccc}
\toprule[2pt]
\multicolumn{5}{c}{WN18} & \multicolumn{5}{c}{FB15k} \\
Model & Methods & CovGap $\downarrow$ & AveSize $\downarrow$ & EF $\downarrow$ & Model & Methods & CovGap $\downarrow$ & AveSize $\downarrow$ & EF $\downarrow$ \\\midrule[1.5pt]

\multirow{6}{*}{TransE} & \textit{KGCP} & \textit{0.096$\pm$0.002}  & \textit{132.36$\pm$6.88} & -- & \multirow{6}{*}{TransE} & \textit{KGCP} &  \textit{0.131$\pm$0.001} & \textit{373.83$\pm$2.08} & -- \\
 & {\MCP} & 0.017$\pm$0.001 & 713.50$\pm$180.91 & 73.56 &  & {\MCP} & 0.021$\pm$0.000 & 583.09$\pm$10.09 & 19.02\\
 & {\ClusterCP}  & 0.073$\pm$0.007 & 117.77$\pm$8.12 & {\ul -6.34} &  & {\ClusterCP} & 0.130$\pm$0.000 & 379.95$\pm$2.36 & 61.20\\
 & {\APS} & 0.108$\pm$0.001 & 11428.73$\pm$817.03 & -- &  & {\APS} & 0.154$\pm$0.001 & 1922.92$\pm$25.04 & -- \\
 & {\RAPS}  & 0.069$\pm$0.001 & 42.03$\pm$1.03 & -- &  & {\RAPS} & 0.124$\pm$0.001 & 336.46$\pm$1.10 & \textbf{-53.39}\\
 & \cellcolor{gray!20} {\CondKGCP} & \cellcolor{gray!20} 0.030$\pm$0.001 & \cellcolor{gray!20} 19.56$\pm$0.14 & \cellcolor{gray!20} \textbf{-17.09} &  & \cellcolor{gray!20} {\CondKGCP} & \cellcolor{gray!20} 0.027$\pm$0.000 & \cellcolor{gray!20} 78.12$\pm$1.23 & \cellcolor{gray!20} {\ul -28.43}\\\midrule

\multirow{6}{*}{RotatE} &  \textit{KGCP} &  \textit{0.076$\pm$0.001} &  \textit{2.13$\pm$0.25} & -- & \multirow{6}{*}{RotatE} &  \textit{KGCP} &  \textit{0.113$\pm$0.001} &  \textit{139.57$\pm$3.44} & --\\
 & {\MCP} & 0.022$\pm$0.003 & 1193.81$\pm$382.91 & 220.68&  & {\MCP} & 0.023$\pm$0.000 & 633.68$\pm$9.80 & {\ul 54.90}\\
 & {\ClusterCP} & 0.056$\pm$0.001 & 1.81$\pm$0.40 & \textbf{-0.16} &  & {\ClusterCP} & 0.113$\pm$0.001 & 141.56$\pm$3.52 & --\\
 & {\APS} & 0.083$\pm$0.002 & 15963.85$\pm$261.07 & -- &  & {\APS} & 0.128$\pm$0.001 & 1757.13$\pm$18.59 & -- \\
 & {\RAPS}  & 0.078$\pm$0.001 & 81.96$\pm$2.13 & -- &  & {\RAPS} & 0.122$\pm$0.000 & 416.41$\pm$3.46 & -- \\
 & \cellcolor{gray!20} {\CondKGCP} & \cellcolor{gray!20} 0.045$\pm$0.001 & \cellcolor{gray!20} 2.20$\pm$0.61 & \cellcolor{gray!20} {\ul 0.02}&  & \cellcolor{gray!20} {\CondKGCP} & \cellcolor{gray!20} 0.063$\pm$0.000 & \cellcolor{gray!20} 246.80$\pm$2.46 & \cellcolor{gray!20} \textbf{21.45}\\\midrule

\multirow{6}{*}{RESCAL} &  \textit{KGCP}  &  \textit{0.103$\pm$0.004} &  \textit{2.61$\pm$0.15} & -- & \multirow{6}{*}{RESCAL} &  \textit{KGCP} &  \textit{0.092$\pm$0.000} &  \textit{62.10$\pm$0.54} & --\\
 & {\MCP} & 0.019$\pm$0.003 & 508.71$\pm$118.00 & 60.25 &  & {\MCP} & 0.020$\pm$0.001 & 740.60$\pm$30.77 & 94.24\\
 & {\ClusterCP} & 0.089$\pm$0.009 & 3.29$\pm$0.60 & {\ul 0.49} &  & {\ClusterCP} & 0.091$\pm$0.001 & 62.25$\pm$0.52 & \textbf{1.50} \\
 & {\APS}  & 0.106$\pm$0.002 & 1298.86$\pm$109.82 & -- &  & {\APS} & 0.085$\pm$0.002 & 369.19$\pm$21.92 & 483.70 \\
 & {\RAPS}  & 0.074$\pm$0.001 & 43.76$\pm$0.46 & 14.19 &  & {\RAPS} & 0.122$\pm$0.000 & 393.44$\pm$1.12 & --  \\
 & \cellcolor{gray!20} {\CondKGCP} & \cellcolor{gray!20} 0.061$\pm$0.001 & \cellcolor{gray!20} 3.80$\pm$0.53 & \cellcolor{gray!20} \textbf{0.28} &  & \cellcolor{gray!20} {\CondKGCP} & \cellcolor{gray!20} {\ul 0.025$\pm$0.000} & \cellcolor{gray!20} 107.91$\pm$0.56 & \cellcolor{gray!20} {\ul 6.84} \\\midrule

\multirow{6}{*}{DistMult} &  \textit{KGCP}  &  \textit{0.066$\pm$0.001} &  \textit{2.30$\pm$0.05} & -- & \multirow{6}{*}{DistMult} &  \textit{KGCP} &  \textit{0.103$\pm$0.001} &  \textit{25.16$\pm$0.12} & --\\
 & {\MCP} & 0.022$\pm$0.002 & 655.33$\pm$135.46 & 148.42 &  & {\MCP} & 0.023$\pm$0.001 & 668.83$\pm$10.05 & 80.46 \\
 & {\ClusterCP} & 0.066$\pm$0.001 & 2.32$\pm$0.05 & -- &  & {\ClusterCP} & 0.103$\pm$0.001 & 25.60$\pm$0.16 & --\\
 & {\APS} & 0.043$\pm$0.002 & 204.67$\pm$39.44 & {\ul 87.99} &  & {\APS} & 0.063$\pm$0.000 & 84.82$\pm$1.37 & {\ul 14.92} \\
 & {\RAPS}  & 0.065$\pm$0.003 & 51.59$\pm$1.04 & 492.90 &  & {\RAPS} & 0.124$\pm$0.000 & 365.76$\pm$0.75 & -- \\
 & \cellcolor{gray!20} {\CondKGCP} & \cellcolor{gray!20} 0.037$\pm$0.001 & \cellcolor{gray!20} 6.18$\pm$0.08 & \cellcolor{gray!20} \textbf{1.34} &  & \cellcolor{gray!20} {\CondKGCP} & \cellcolor{gray!20} 0.024$\pm$0.000 & \cellcolor{gray!20} 66.27$\pm$0.96 & \cellcolor{gray!20} \textbf{5.20} \\\midrule

\multirow{6}{*}{ComplEx} &  \textit{KGCP}  &  \textit{0.072$\pm$0.001} &  \textit{1.07$\pm$0.01} & -- & \multirow{6}{*}{ComplEx} &  \textit{KGCP} &  \textit{0.088$\pm$0.001} &  \textit{34.99$\pm$0.88} & -- \\
 & {\MCP} & 0.023$\pm$0.002 & 1898.69$\pm$226.32  & {\ul 387.27} &  & {\MCP} & 0.024$\pm$0.002 & 664.43$\pm$19.46 & 98.35 \\
 & {\ClusterCP} & 0.072$\pm$0.001 & 1.07$\pm$0.01 & -- &  & {\ClusterCP} & 0.088$\pm$0.001 & 34.88$\pm$0.85 & -- \\
 & {\APS} & 0.065$\pm$0.004 & 15669.60$\pm$498.74 & 22383.61 &  & {\APS} & 0.054$\pm$0.001 & 177.94$\pm$10.72 & {\ul 42.04} \\
 & {\RAPS} & 0.074$\pm$0.004 & 63.82$\pm$2.43 & -- &  & {\RAPS} & 0.121$\pm$0.000 & 417.50$\pm$3.41 & --\\
 & \cellcolor{gray!20} {\CondKGCP} & \cellcolor{gray!20} 0.049$\pm$0.002 & \cellcolor{gray!20} 1.39$\pm$0.01 & \cellcolor{gray!20} \textbf{0.14} &  & \cellcolor{gray!20} {\CondKGCP} & \cellcolor{gray!20} 0.026$\pm$0.000 & \cellcolor{gray!20} 166.10$\pm$5.23 & \cellcolor{gray!20} \textbf{21.15} \\\midrule

\multirow{6}{*}{ConvE} &  \textit{KGCP}  &  \textit{0.066$\pm$0.004} &  \textit{1.71$\pm$0.04} & -- & \multirow{6}{*}{ConvE} &  \textit{KGCP} &  \textit{0.102$\pm$0.001} &  \textit{91.02$\pm$7.06} & --\\
 & {\MCP} & 0.019$\pm$0.002 & 576.97$\pm$170.39 & {\ul 122.40} &  & {\MCP} & 0.023$\pm$0.000 & 725.08$\pm$38.15 & {\ul 80.26} \\
 & {\ClusterCP} & 0.066$\pm$0.001 & 1.72$\pm$0.04 & -- &  & {\ClusterCP} & 0.102$\pm$0.001 & 88.33$\pm$7.06 & --\\
 & {\APS} & 0.071$\pm$0.004 & 9.85$\pm$1.77 & -- &  & {\APS} & 0.110$\pm$0.003 & 4578.40$\pm$243.77 & -- \\
 & {\RAPS} & 0.068$\pm$0.002 & 47.75$\pm$1.44 & -- &  & {\RAPS} & 0.123$\pm$0.001 & 400.08$\pm$4.54 & -- \\
 & \cellcolor{gray!20} {\CondKGCP} & \cellcolor{gray!20} 0.038$\pm$0.001 & \cellcolor{gray!20} 4.80$\pm$0.10 & \cellcolor{gray!20} \textbf{1.10} &  & \cellcolor{gray!20} {\CondKGCP} & \cellcolor{gray!20} 0.032$\pm$0.000 & \cellcolor{gray!20} 429.14$\pm$3.11 & \cellcolor{gray!20} \textbf{48.30} \\
\bottomrule[2pt]
\end{tabular}%
}
\caption{Overall performance comparison of {\CondKGCP} and baseline methods across six KGE models and two benchmark datasets (WN18 and FB15k). We report {\CovGap} and {\AvgSize} as the mean $\pm$ standard deviation over 10 independent trials. The EF (efficient rate) is reported as a mean value; its standard deviation is omitted as it is negligible.
The best and second-best EF values for each model-dataset pair are highlighted in bold and {\ul underline}, respectively.
{\KGCP} is shown in italic to indicate that it serves as a baseline without conditional coverage guarantees. Our proposed method, {\CondKGCP}, is highlighted with a \colorbox{gray!20}{gray} background.
“--” in the EF column denotes a failure case where either {\CovGap} is not reduced or {\AvgSize} does not change relative to {\KGCP}.}
\label{tab:main_filter}
\end{table*}

\section{Experiment}
We evaluate the {\CondKGCP} empirically and demonstrate its effectiveness in balancing the trade-off between predicate-conditional coverage probability and the size of the prediction sets.

\subsection{Experimental Setup}
\textbf{Training KGE Models}. We trained our KGE models using the LibKGE framework \cite{libkge}, following the hyperparameter search strategy described by \citet{ruffinelli2019you}. All experiments were conducted on a Linux machine equipped with a 40GB NVIDIA A100 SXM4 GPU.

\textbf{Datasets}. We consider two widely-used benchmark datasets: WN18 and FB15k \cite{bordes2013translating}. 
%Note that our objective is to evaluate the uncertainty quantification. 
We follow \citet[Appendix D.1]{zhu2024conformalized} and do not consider their modified counterparts, WN18RR \cite{dettmers2018convolutional} and FB15k-237 \cite{toutanova2015observed} since they are not suitable for evaluating uncertainty quantification.
%Their modified counterparts, WN18RR \cite{dettmers2018convolutional} and FB15k-237 \cite{toutanova2015observed}, address test set leakage and increase the difficulty of training and generalization by requiring models to capture more complex patterns.
%However, these modified datasets are not suitable for evaluating uncertainty quantification, as discussed in \citet[Appendix D.1]{zhu2024conformalized}.

\textbf{Baselines}. We consider following methods as baselines: (1) {\KGCP} \cite{zhu2024conformalized}; (2) {\MCP} \cite{vovk2005algorithmic} which performs conformal prediction at the predicate-level; (3) {\ClusterCP} \cite{ding2024class}, which clusters predicates based on similarity of score distribution and then conducts conformal prediction at the cluster level; (4) {\APS} \cite{romano2020classification} and {\RAPS} \cite{Angelopoulos2021image}, which modify the nonconformity measure to provide improved conditional coverage probabilities (see details in Appendix \ref{app:advanced_scores}). 
Unless otherwise specified, we use the default nonconformity measure, {\Softmax}, as proposed by \citet{zhu2024conformalized} (see the definition in Appendix \ref{app:default_score}). 
% The conformal prediction framework follows \citet{zhu2024conformalized}. For the nonconformity score, we used {\Softmax} as the primary nonconformity measure in Table \ref{tab:main_filter}. Additionally, we conducted sensitivity tests comparing various nonconformity measures: $NegScore$, {\Softmax}, and $Minmax$ from \citet{zhu2024conformalized}, as well as {\APS} from \citet{romano2020classification} and {\RAPS} from \citet{Angelopoulos2021image}. Results for the sensitivity tests are presented in Table \ref{tab:score}, with further details in Appendix \ref{app:nonconf}.

\textbf{Hyperparameter Tuning}. 
The validation set serves as the calibration set for all baselines. For {\ClusterCP}, {\RAPS} and {\CondKGCP}, we randomly sample triples from the training set (of the same sizes as the calibration set) to determine optimal hyperparameter settings.
We follow hyperparameter search strategy from the original papers for {\ClusterCP} and {\RAPS}. 
For {\CondKGCP}, we tune $\gamma\in [0.01, 0.1, 0.5]$ and $\phi\in [20,50,100,200]$. The best hyperparameters are reported in Table~\ref{tab:condkgcp_hyper}.

\subsection{Evaluation Setup}
We set the target coverage probability $1-\epsilon=0.9$ by default, following \citet{zhu2024conformalized}. 
For each KGE method-dataset pair, we train the model 10 times, each time using a different random seed, and report the mean and standard deviation.
%Experiments are repeated over 10 times by training KGE models with different random seeds, and we report the mean and standard deviation.

\textbf{Evaluation Metrics.} Following \citet{ding2024class}, we evaluate the performance using two metrics: coverage gap ({\CovGap}) and average size of the prediction sets ({\AvgSize}).

Given a set of test triples $\TestSet$, the empirical predicate-conditional coverage for each predicate $r\in R$, denoted as $\CovR$, is calculated as:
\begin{equation}
    \CovR = \frac{1}{|\TestSet[\{r\}]|}\sum_{(q,e)\in\TestSet[\{r\}]}\mathbbm{1}[e\in \hat C(q)].
\end{equation}
%Given a user-specified error rate $\epsilon$, we measure how closely the empirical predicate-conditional coverage aligns with the desired coverage ratio $1 - \epsilon$. To do this, we define the average predicate coverage gap ({\CovGap}) as:
The average predicate-conditional coverage gap ({\CovGap}) measures how far the empirical coverage is from the desired coverage level $1-\epsilon$:
%is defined as:
\begin{equation}
    \operatorname{CovGap} = \frac{1}{|R|}\sum_{r\in R}|\CovR - (1-\epsilon)|
\end{equation}
%In addition to coverage, we evaluate the efficiency of the predictors by measuring the size of the answer sets. The average size of the prediction sets ({\AvgSize}) is computed as:
The average size of the prediction sets ({\AvgSize}) is computed as 
\begin{equation}
    \frac{1}{\TestSet[\{r\}]}\sum_{(q,e)\in\TestSet[\{r\}]}|\hat C(q)|.
\end{equation}

Note that {\CovGap} and {\AvgSize} are inherently competing metrics in conformal prediction \cite{Angelopoulos2021image}; reducing {\CovGap} often increases {\AvgSize}. For a given {\CovGap}, smaller {\AvgSize} is preferred for more informative estimates.
Rather than optimizing either metric in isolation, our aim is to balance the trade-off between coverage probability and prediction set size.
To quantify this trade-off, we introduce an auxiliary metric \emph{Efficiency Rate} (ER): the number of additional entities required (relative to KGCP) to reduce {\CovGap} by 0.01:
\begin{equation}
    \frac{\operatorname{AvgSize}-\operatorname{AvgSize}^*}{\operatorname{CovGap}^*-\operatorname{CovGap}}\times 0.01,
\end{equation}
where $(\operatorname{AvgSize}^*, \operatorname{CovGap}^*)$ are the corresponding values for KGCP.
%As a result, it is hard to determine whether method A is better than method B when method A achieves a lower {\CovGap} but a higher {\AvgSize}.
% To balance {\CovGap} and {\AvgSize}, we define trade-off ratio ({\TOR}) as:
% \begin{equation}
%     \operatorname{TOR} = \frac{\operatorname{CovGap}\cdot\operatorname{AveSize}}{\operatorname{CovGap}^*\cdot\operatorname{AveSize}^*},
% \end{equation}
% where $\operatorname{CovGap}^*$ and $\operatorname{AveSize}^*$ are {\CovGap} and {\AvgSize} for the baseline method {\KGCP}.

% Under coverage ratio ($\operatorname{UCR}$) is defined as
% \begin{equation}
%     \operatorname{UCR} := \frac{1}{|R|}\sum_{r\in R}\mathbbm{1}\{Cov_r<1-\epsilon\}
% \end{equation}

\subsection{Results and Discussion}
\subsubsection{Overall Comparison}
Table \ref{tab:main_filter} presents a comprehensive comparison of {\CondKGCP} with baseline methods across six KGE models and two benchmark datasets (WN18 and FB15k). 
\textbf{Overall, {\CondKGCP} consistently demonstrates the most favorable trade-off between coverage and prediction set size, as evidenced by its lowest EF scores in the majority of cases.}
%The results demonstrate that {\CondKGCP} achieves {\CovGap} values close to the empirical lower bound across all evaluated KGE models and datasets, while maintaining compact prediction sets.
%consistently achieves the best trade-off between {\CovGap} and {\AvgSize}, as reflected in its superior {\TOR} values.

Among the methods, {\MCP} achieves the lowest {\CovGap}, aligning with Proposition 4.6 in \citet{vovk2005algorithmic}.
Although {\CovGap} is not exactly zero due to the small number of triples for certain subgroups, this deviation is minimal.
Thus, {\MCP}'s {\CovGap} can be viewed as the empirical lower bound.
However, this improved coverage precision comes at a significant cost: {\MCP} produces substantially larger prediction sets, resulting in high {\AvgSize}. 
In contrast, {\KGCP} is designed to achieve only marginal coverage guarantees, resulting in higher {\CovGap}. However, this enables {\KGCP} to generate the smallest prediction sets in most cases, yielding the lowest {\AvgSize} empirically. 
We observe that \textbf{{\CondKGCP} achieves {\CovGap} values that are consistently closest to the empirical lower bound, while maintaining compact prediction sets}--often with {\AvgSize} values closest to the empirical lower bound-compared to other baseline methods across all evaluated KGE methods and datasets.

Note that while {\ClusterCP} occasionally achieves lower {\AvgSize} values comparable to the empirical lower bound, it fails to reduce {\CovGap} in these cases.
In fact, both its {\CovGap} and {\AvgSize} values remain very close to those of {\KGCP}, the baseline method with marginal coverage guarantees (i.e., calibrated on triples across all predicates), indicating that {\ClusterCP} fails to cluster predicates into meaningful groups.
This limitation arises because {\ClusterCP} is designed for subgroups with similar data points, making it unsuitable for our setting, where the distribution of triples across predicates is highly imbalanced.

%Improving conditional coverage typically increases the size of prediction sets.
Methods that modify the nonconformity score, such as {\APS} and {\RAPS}, often generate overly conservative prediction sets due to prioritization on difficult regions for coverage.
As shown in Table \ref{tab:main_filter}, neither {\APS} nor {\RAPS} significantly improve {\CovGap} and frequently generate large prediction sets, suggesting the nonconformity measures might not be well-suited for KGE methods.
%{\ClusterCP} intend to improve {\CovGap} by clustering predicates into meaningful groups. However, the coarse granularity of the predicate clusters often leads to performance similar to {\KGCP}, as observed in our results.
%Subgroup-based methods ({\MCP}, {\ClusterCP}, {\CondKGCP}) inherently produce larger prediction sets due to the reduced calibration data within subgroups, which lowers calibration precision. 
%{\ClusterCP} offers improvements of {\CovGap} by clustering predicates into meaningful groups. However, the coarse granularity of the predicate clusters often leads to performance similar to {\KGCP}, as observed in our results. 

\subsubsection{Comparison across Different Coverage Levels} 
\begin{figure}[t!]
    \centering
    \includegraphics[width=\linewidth]{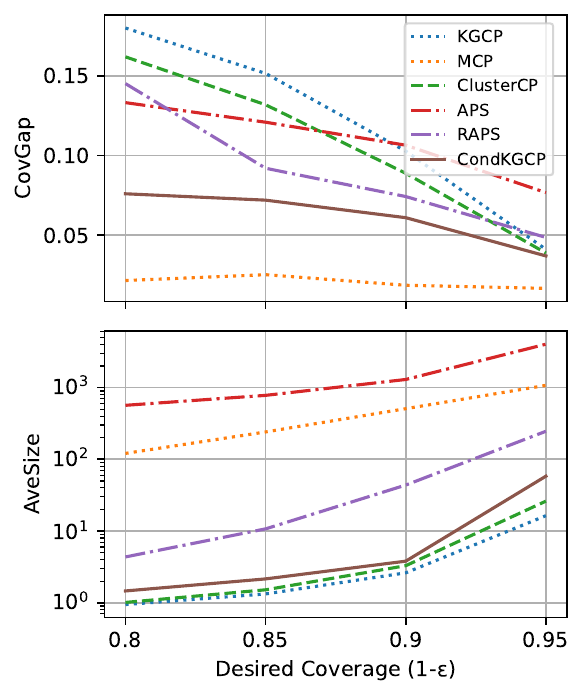}
    \caption{Comparison of methods across varying target coverage levels, showing {\CovGap} (top plot) and {\AvgSize} (bottom plot) for RESCAL on WN18. Complete results are provided in Tables \ref{fig:alpha_covgap} and \ref{fig:alpha_size} in the Appendix. }
    \label{fig:alpha_inves}
\end{figure}

To evaluate performance under varying coverage levels, we conduct experiments for different target coverage probability $1 - \epsilon\in [0.8, 0.85, 0.9, 0.95]$. The results, shown in Figure \ref{fig:alpha_inves}, reveal that {\CondKGCP} achieves {\CovGap} values closest to the empirical lower bound ({\MCP}) while maintaining {\AvgSize} comparable to the empirical lower bound ({\KGCP}) across all coverage levels. 

\begin{table*}[t!]
\centering
\resizebox{\textwidth}{!}{%
\begin{tabular}{ccccc||ccccc}
\toprule[2pt]
\multicolumn{5}{c}{WN18} & \multicolumn{5}{c}{FB15k} \\
Model & Method & CovGap $\downarrow$ & AveSize $\downarrow$ & EF $\downarrow$ & Model & Method & CovGap $\downarrow$ & AveSize $\downarrow$ & EF $\downarrow$ \\\midrule
\multirow{3}{*}{TransE} & w/o Merge & 0.097$\pm$0.002 & 139.54$\pm$6.50 & -- & \multirow{3}{*}{TransE} & w/o Merge & 0.131$\pm$0.002 & 371.89$\pm$4.11 & -- \\
 & w/o RankCal & 0.019$\pm$0.001 & 629.13$\pm$129.55 & 791.79 &  & w/o RankCal & 0.022$\pm$0.000 & 98.24$\pm$2.14 & -25.28\\
 & {\CondKGCP} & 0.030$\pm$0.001 & 19.56$\pm$0.14 & \textbf{-17.09}&  & {\CondKGCP} & 0.027$\pm$0.000 & 78.12$\pm$1.23 & \textbf{-28.43}\\\midrule
 
\multirow{3}{*}{RotatE} & w/o Merge & 0.077$\pm$0.002 & 1.98$\pm$0.54 & -- & \multirow{3}{*}{RotatE} & w/o Merge & 0.114$\pm$0.001 & 141.79$\pm$1.32 & -- \\
 & w/o RankCal & 0.023$\pm$0.001 & 1781.28$\pm$230.30 & 335.69 &  & w/o RankCal & 0.043$\pm$0.000 & 370.29$\pm$3.55 & 32.96\\
 & {\CondKGCP} & 0.045$\pm$0.001 & 2.20$\pm$0.61 & \textbf{0.02} &  & {\CondKGCP} & 0.063$\pm$0.000 & 246.80$\pm$2.46 & \textbf{21.45}\\\midrule
 
\multirow{3}{*}{RESCAL} & w/o Merge & 0.105$\pm$0.003 & 2.57$\pm$0.40 & -- & \multirow{3}{*}{RESCAL} & w/o Merge & 0.090$\pm$0.002 & 62.55$\pm$0.25 & \textbf{2.25} \\
 & w/o RankCal & 0.021$\pm$0.000 & 385.33$\pm$63.64 & 46.67 &  & w/o RankCal & 0.020$\pm$0.000 & 134.22$\pm$0.88 & 10.02\\
 & {\CondKGCP} & 0.061$\pm$0.001 & 3.80$\pm$0.53 & \textbf{0.28} &  & {\CondKGCP} & 0.025$\pm$0.000 & 107.91$\pm$0.56 & 6.84\\\midrule
 
\multirow{3}{*}{DistMult} & w/o Merge & 0.066$\pm$0.001 & 3.39$\pm$0.04 & -- & \multirow{3}{*}{DistMult} & w/o Merge & 0.103$\pm$0.002 & 25.82$\pm$0.21 & -- \\
 & w/o RankCal & 0.023$\pm$0.000 & 719.51$\pm$111.22 & 166.79 &  & w/o RankCal & 0.023$\pm$0.000 & 98.00$\pm$1.22 & 9.11\\
 & {\CondKGCP} & 0.037$\pm$0.001 & 6.18$\pm$0.08 & \textbf{1.34} &  & {\CondKGCP} & 0.024$\pm$0.000 & 66.27$\pm$0.96 & \textbf{5.20}\\\midrule
 
\multirow{3}{*}{ComplEx} & w/o Merge & 0.074$\pm$0.002 & 1.07$\pm$0.01 & -- & \multirow{3}{*}{ComplEx} & w/o Merge & 0.086$\pm$0.001 & 56.49$\pm$0.44  & 107.50 \\
 & w/o RankCal & 0.025$\pm$0.000 & 2313.94$\pm$260.99 & 492.10&  & w/o RankCal & 0.026$\pm$0.000 & 168.50$\pm$5.10  & 21.53\\
 & {\CondKGCP} & 0.049$\pm$0.002 & 1.39$\pm$0.01 & \textbf{0.14} &  & {\CondKGCP} & 0.026$\pm$0.000 & 166.10$\pm$5.23 & \textbf{21.15}\\\midrule
 
\multirow{3}{*}{ConvE} & w/o Merge & 0.066$\pm$0.001 & 2.74$\pm$0.05 & -- & \multirow{3}{*}{ConvE} & w/o Merge & 0.099$\pm$0.002 & 180.79$\pm$2.79 & 299.23\\
 & w/o RankCal & 0.021$\pm$0.000 & 575.14$\pm$89.71 & 127.43 &  & w/o RankCal & 0.027$\pm$0.000 & 579.14$\pm$5.03 & 65.08\\
 & {\CondKGCP} & 0.038$\pm$0.001 & 4.80$\pm$0.10 & \textbf{1.10} &  & {\CondKGCP} & 0.032$\pm$0.000 & 429.14$\pm$3.11 & \textbf{48.30}\\
 \bottomrule[2pt]
\end{tabular}%
}
\caption{Ablation Study of {\CondKGCP}. Each model is evaluated under three configurations: without predicate merging procedure (w/o Merge), without rank calibration (w/o RankCal), and the proposed {\CondKGCP} (full method). }
\label{tab:ablation}
\end{table*}

\subsubsection{Impact of Hyperparameters} 
{\CondKGCP} includes two key hyperparameters: $\phi$, which controls the granularity of predicate subgrouping in the merging procedure, and $\gamma$, which balances the conditional coverage guarantee and the size of prediction sets in the dual calibration schema. 

\begin{figure}[t!]
    \centering
    \includegraphics[width=\linewidth]{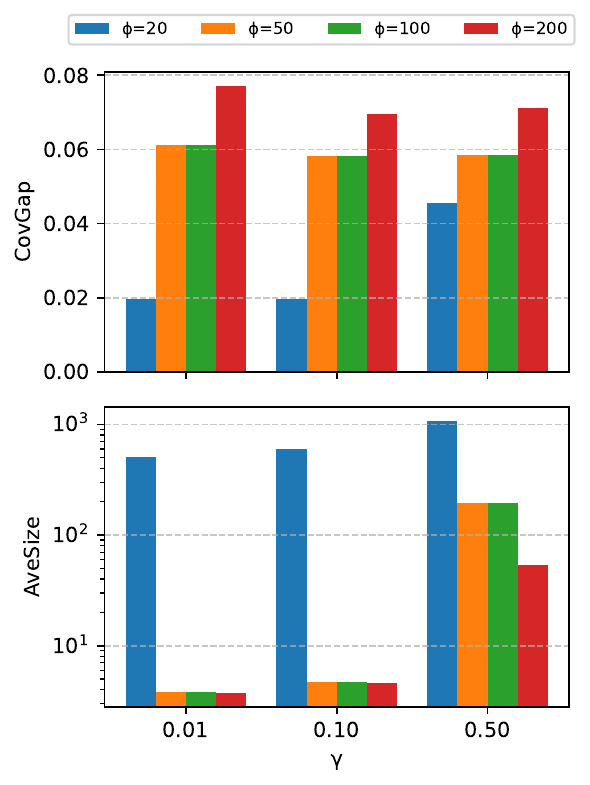}
    \caption{Influence of hyperparameters $\phi$ and $\gamma$ on {\CovGap} (top) and {\AvgSize} (bottom) for RESCAL on WN18. Complete results for all model-dataset combinations are provided in Tables \ref{fig:hyper_covgap} and \ref{fig:hyper_size} in the Appendix. }
    \label{fig:hyper_inves}
\end{figure}

As shown in Figure \ref{fig:hyper_inves}, larger values of $\phi$ result in increased {\CovGap} and reduced {\AvgSize}. This is expected, as coarser subgrouping theoretically results in behavior more similar to {\KGCP}, whereas finer subgrouping aligns more closely with {\MCP}.

For $\gamma$, larger values are associated with increased {\AvgSize}, consistent with the analysis in section 5. However, the empirical results reveal a key practical insight: adjusting $\gamma$ to sacrifice a small amount in the lower bound of conditional coverage can significantly reduce {\AvgSize} while causing only a negligible change in {\CovGap}. This demonstrates the necessity of including $\gamma$ in the design of {\CondKGCP}.

\subsubsection{Impact of Nonconformity Measure}
Subgroup-based methods ({\MCP}, {\ClusterCP}, {\CondKGCP}) can be combined with nonconformity score-based methods ({\APS}, {\RAPS}). However, as shown in Table \ref{tab:score_results} in the Appendix, these combinations do not result in a better trade-off. 
Notably, {\CondKGCP} achieves low {\CovGap} values with significantly smaller prediction sets in most cases compared to {\MCP} and {\ClusterCP}, regardless of the choice of the nonconformity measure. This observation demonstrates that the improvements offered by {\CondKGCP} are robust to the change of the nonconformity measure.

\subsubsection{Impact of Key Components}
We compare {\CondKGCP} with two variants: {\CondKGCP} without predicate merging procedure (w/o Merge) and {\CondKGCP} without rank calibration (w/o RankCal). 
The results in Table~\ref{tab:ablation} show that {\CondKGCP} outperforms both variants in balancing {\CovGap} and {\AvgSize}, achieving the lowest EF and thus the most efficient trade-off.
%Notably, rank calibration is particularly relevant in KGs with fewer predicates, as evidenced by the results for WN18, where the {\TOR} of w/o RankCal is much higher than the other variants.
Concretely, {\CondKGCP} achieves lower {\CovGap} compared to w/o Merge with a comparable {\AvgSize}. While it shows slightly higher {\CovGap} than w/o RankCal, it maintains smaller {\AvgSize}.
This demonstrates the contribution of each component: the merging process effectively reduces {\CovGap}, while rank calibration ensures more compact prediction set. 

\section{Discussion and Conclusion}
In this paper, we introduces {\CondKGCP}, a novel method that addresses the limitations of existing conformalized KGE method by approximating predicate-conditional coverage guarantees while maintaining compact prediction sets. 
We theoretically prove that the deviation from the desired confidence level is bounded and empirically demonstrate the effectiveness of {\CondKGCP} across six KGE methods and two benchmark datasets.

Our method offers a useful uncertainty quantification tool for high-stakes applications, such as medical diagnosis, and can be easily adapted to quantify uncertainty under other types of conditions, such as entity-type. Additionally, it can be seamlessly extended to other tasks, including embedding-based query answering \cite{ren2020query2box} and probabilistic reasoning over KG \cite{zhu2023towards, zhu2024approximating}.

%Our approximation provides a useful conditional uncertainty quantification tool in practice. 
%That is, given a desired confidence level e.g. 90\%, the average deviation to the confidence level is around 4\% and 3\% for WN18 and FB15k with reasonably small prediction sets 6.3/40,943 for WN18 and 182.4/14,951 for FB15k.
%By merging predicates with similar vector representations and incorporating a dual calibration schema that leverages both score and rank information, {\CondKGCP} balances the trade-off between reliable uncertainty quantification and prediction set size. 

%This method can be easily adapted to other conditional uncertainty quantification application such as entity type conditional uncertainty quantification.
%Theoretical guarantees demonstrate its validity, and extensive empirical evaluations on benchmark datasets reveal its superiority over baseline methods in terms of conditional coverage and prediction set size. 

\section{Limitation}
A potential limitation of the proposed {\CondKGCP} lies in the probabilistic guarantees provided by Proposition \ref{prop:cov}, which rely on the assumption of i.i.d. (or weaker exchangeability \cite{vovk2005algorithmic}) data, as well as the assumption that the similarity of vector representations corresponds to the similarity of the distribution of nonconformity scores. 
While the i.i.d. assumption may occasionally be violated in certain real-world applications, it is a common simplification in statistical methods. As a step forward, we are working on extending our approach to handle covariate shift, where only the input distribution changes while the conditional distribution remains unchanged.

Regarding the similarity assumption, the effectiveness of the predicate merging step, as demonstrated in the ablation study (Table \ref{tab:ablation}), indicates that this assumption is reasonable for KGE methods. Nonetheless, future work could explore incorporating additional features, such as the semantic meaning of predicates, to further enhance the merging process and improve robustness in diverse scenarios.

\section{Acknowledgments}
The authors thank the International Max Planck Research School for Intelligent Systems (IMPRS-IS) for supporting Yuqicheng Zhu. The work was partially supported by EU Projects Graph Massivizer (GA 101093202), enRichMyData (GA 101070284) and SMARTY (GA 101140087), as well as the Deutsche Forschungsgemeinschaft (DFG, German Research Foundation) – SFB 1574 – 471687386. Zifeng Ding receives funding from the European Research Council (ERC) under the European Union’s Horizon 2020 Research and Innovation programme grant AVeriTeC (Grant agreement No. 865958).

\newpage

\bibliography{anthology}

\appendix

\onecolumn
\section{Proofs}\label{app:proof}
\setcounter{proposition}{0}
\setcounter{corollary}{1}

\begin{proposition}[Conditional Coverage Guarantee]
% Given a user-specified error rate $\epsilon$ and a $\gamma\in[0,1]$, the conditional coverage probability of {\CondKGCP} has the following bounds for all $g\in\{1,2, \dots, G\}$ if there is no ties between scores in $\{S(tr_i)\}_{i\in\mathcal{I}^g}$:
% \begin{equation}
%     \mathbb{P}(e_{test}\in\hat C(q_{test})|f(Pre(q_{test}))=g)\in [1-\epsilon-(1-\gamma)\epsilon_g^{\hat k(g)}, 1-\epsilon+\gamma\epsilon_g^{\hat k(g)}+\frac{1}{|\mathcal{I}^g|+1}]
% \end{equation}
Let $q$ be a query and $e$ be its answer entity.
Given a user-specified error rate $\epsilon$ and a $\gamma\in[0,1]$, we have the following bounds for all parts $g \in P$:
\begin{align}
    \mathbb{P}(e\in \PredicitonSetCondKGCP(q)\mid \pred{q} \in g) \geq 1-\epsilon-(1-\gamma)\epsilon_g^{\hat k(g)},
\end{align}
and if there is no tie in the set of nonconformity scores of the triples in $\PartSet$, then
\begin{align}
    \mathbb{P}(e\in \PredicitonSetCondKGCP(q)\mid \pred{q} \in g) \leq 1-\epsilon+\gamma\epsilon_g^{\hat k(g)}+\frac{1}{|\PartSet|+1}.
\end{align}
\end{proposition}

\begin{proof}[Proof of the lower bound]
    We prove the lower bound similar to the proof of Theorem 4.1 in \citet[Appendix A.1]{shi2024conformal}. 
    %$g\in \{1,2, \dots, G\}$ denote the index of merged predicate subgroup, 
    We first conduct conformal prediction for each part $g\in P$ (which can be viewed as applying {\MCP} where each subgroup contains triples whose predicates are in a predicate set generated by Algorithm \ref{alg:set-partition}) with the adjust error rate $\epsilon'(g)$. The prediction set is defined as follows:
    \begin{equation}
        \hat C_{\operatorname{MCP^*}}(q) = \FilteredE{S}{\hat s_{\epsilon'(g)}(\PartSet)}
    \end{equation}
    % where 
    % \begin{equation}\label{eq:answerset}
    %      \hat s^g_{\epsilon'} = \Quantile(\frac{\lceil(|\mathcal{I}^g|+1)(1-\epsilon')\rceil}{|\mathcal{I}^g|}, \{S(tr_i)\}_{i\in\mathcal{I}^g}),
    % \end{equation}
    According to \cite[Proposition 4.6]{vovk2005algorithmic}, for a query $q$, we have
    \begin{equation}
        \mathbb{P}(e\in\hat C_{\operatorname{MCP^*}}(q)\mid\pred{q}\in g)\geq 1-\epsilon'
    \end{equation}
    Suppose that the rank threshold for part $g$ is $\RankThreshold$ and $\epsilon_g^{\hat k(g)}$ is its corresponding miscoverage error of top-$\RankThreshold$ prediction sets, we have
    % \begin{align*}
    %     &\mathbb{P}(e_{test}\in\hat C_{MCP}(q_{test})|f(Pre(q_{test}))=g)\\
    %     =&\mathbb{P}(S(tr(q_{test}, e_{test}))\leq\hat s^g_{\epsilon'}|f(Pre(q_{test}))=g)\\
    %     =&\mathbb{P}(S(tr(q_{test}, e_{test}))\leq\hat s^g_{\epsilon'}, r_{M_\theta}(q_{test}, e_{test})\leq\hat k(g)|f(Pre(q_{test}))=g)\\
    %     &+ \mathbb{P}(S(tr(q_{test}, e_{test}))\leq\hat s^g_{\epsilon'}, r_{M_\theta}(q_{test}, e_{test})>\hat k(g)|f(Pre(q_{test}))=g)\\
    %     \leq&\mathbb{P}(S(tr(q_{test}, e_{test}))\leq\hat s^g_{\epsilon'}, r_{M_\theta}(q_{test}, e_{test})\leq\hat k(g)|f(Pre(q_{test}))=g)\\
    %     &+\underbrace{\mathbb{P}(r_{M_\theta}(q_{test}, e_{test})>\hat k(g)|f(Pre(q_{test}))=g)}_{\epsilon_g^{\hat k(g)}}
    % \end{align*}
    \begin{align*}
        &\mathbb{P}(e\in\hat C_{\operatorname{MCP^*}}(q)\mid\pred{q}\in g)\\
        =&\mathbb{P}\Big(S(tr(q, e))\leq\hat s_{\epsilon'(g)}(\PartSet)\mid\pred{q}\in g\Big)\\
        =&\mathbb{P}\Big(S(tr(q, e))\leq\hat s_{\epsilon'(g)}(\PartSet), \AnswerRank\leq\RankThreshold\mid\pred{q}\in g\Big)\\
        &+\mathbb{P}\Big(S(tr(q, e))\leq\hat s_{\epsilon'(g)}(\PartSet), \AnswerRank>\RankThreshold\mid\pred{q}\in g\Big)\\
        \leq&\mathbb{P}\Big(S(tr(q, e))\leq\hat s_{\epsilon'(g)}(\PartSet), \AnswerRank\leq\RankThreshold\mid\pred{q}\in g\Big)\\
        &+\underbrace{\mathbb{P}(\AnswerRank>\RankThreshold\mid\pred{q}\in g)}_{\RankThreshold}\\
    \end{align*}
    By definition of the prediction set constructed by {\CondKGCP}, we have
    \begin{equation}
        \mathbb{P}(e\in\PredicitonSetCondKGCP(q)\mid \pred{q}\in g)\geq 1-\epsilon'(g)-\epsilon_g^{\hat k(g)}
    \end{equation}
    In our paper we set $\epsilon'(g)=\epsilon-\gamma\cdot\epsilon_g^{\hat k(g)}$, therefore, we have
    \begin{equation}
        \mathbb{P}(e\in\PredicitonSetCondKGCP(q)\mid \pred{q}\in g)\geq 1-\epsilon-(1-\gamma)\epsilon_g^{\hat k(g)}
    \end{equation}
\end{proof}

\begin{proof}[Proof of the upper bound]
    We prove the upper bound based on \citet[Appendix A.1]{lei2018distribution}. By assuming no ties in the set of nonconformity scores of the triples in $\PartSet$, denoted as $\mathcal{S}(\PartSet)$, we know that the  nonconformity scores in $\mathcal{S}(\PartSet)$ are all distinct with probability one.
    The set $\hat C_{\operatorname{MCP^*}}(q)$ is equivalent to the set of all answer entity $e'\in E$ such that the nonconformity score $S(tr(q, e'))$ ranks among the $\lceil(|\PartSet|+1)(1-\epsilon'(g))\rceil$ smallest of $\mathcal{S}(\PartSet)$.
    
    Consider now the complementary set $\hat D_{\operatorname{MCP^*}}(q)$ consisting of answer entities $e'\in E$ such that the nonconformity score $S(tr(q, e'))$ is among the $\lceil(|\PartSet|+1)\epsilon'(g)-1\rceil$ largest. 
    Under i.i.d assumption (or a weaker exchangeability assumption), the joint distribution of the nonconformity scores $\mathcal{S}(\PartSet)$ is invariant under permutations. As a result, the ranks of the nonconformity scores in $\mathcal{S}(\PartSet)$ are uniformly distributed among $\{1, 2, \dots, |\PartSet|\}$ and hence we can derive the following lower bound for each part $g\in P$:
    \begin{equation}
        \mathbb{P}(e\in \hat D_{\operatorname{MCP^*}}(q)\mid\pred{q}\in g)\geq \epsilon'(g)-\frac{1}{|\PartSet|+1},
    \end{equation}
    We also know that there is no intersected elements for $\hat C_{\operatorname{MCP^*}}(q)$ and $\hat D_{\operatorname{MCP^*}}(q)$
    \begin{equation}
        \hat C_{\operatorname{MCP^*}}(q)\cap\hat D_{\operatorname{MCP^*}}(q)=\emptyset
    \end{equation}
    Then we can derive the upper bound for each part $g$ as follows:
    \begin{align*}
        &\mathbb{P}(e\in\hat C_{\operatorname{MCP^*}}(q)\mid\pred{q}\in g) + \mathbb{P}(e\in\hat D_{\operatorname{MCP^*}}(q)\mid\pred{q}\in g)\leq 1\\
        \Rightarrow&\mathbb{P}(e\in\hat C_{\operatorname{MCP^*}}(q)\mid\pred{q}\in g)\leq 1 -  \mathbb{P}(e\in\hat D_{\operatorname{MCP^*}}(q)\mid\pred{q}\in g)\\
        \Rightarrow&\mathbb{P}(e\in\hat C_{\operatorname{MCP^*}}(q)\mid\pred{q}\in g)\leq 1-\epsilon'(g)+ \frac{1}{|\PartSet|+1}\\
        \Rightarrow&\mathbb{P}(e\in\hat C_{\operatorname{MCP^*}}(q)\mid\pred{q}\in g)\leq 1-\epsilon+\gamma\epsilon_g^{\hat k(g)}+\frac{1}{|\mathcal{I}^g|+1}\\
    \end{align*}
    Based on the definition of {\CondKGCP}, we have
    \begin{align*}
        &\mathbb{P}(e\in\hat C_{\operatorname{MCP^*}}(q)\mid\pred{q}\in g)\\
        =&\underbrace{\mathbb{P}\Big(S(tr(q, e))\leq\ScoreThreshold, \AnswerRank\leq\RankThreshold\mid\pred{q}\in g\Big)}_{\mathbb{P}(e\in\PredicitonSetCondKGCP(q)\mid \pred{q}\in g)}\\
        &+ \mathbb{P}\Big(S(tr(q, e))\leq\ScoreThreshold, \AnswerRank>\hat k(g)\mid  \pred{q}\in g\Big)\\
        \Rightarrow& \mathbb{P}(e\in\PredicitonSetCondKGCP(q)\mid \pred{q}\in g)\leq\mathbb{P}(e\in\hat C_{\operatorname{MCP^*}}(q)\mid \pred{q}\in g)\\
        \Rightarrow& \mathbb{P}(e\in\PredicitonSetCondKGCP(q)\mid \pred{q}\in g)\leq 1-\epsilon+\gamma\epsilon_g^{\hat k(g)}+\frac{1}{|\PartSet|+1}
    \end{align*}
    
\end{proof}

\begin{corollary}[\citet{shi2024conformal}]
    % Suppose $\hat\epsilon_g$ and $\hat k(g)$ satisfy 
    % \begin{equation}
    %     \hat k(g)\in\{k:\epsilon_g^k<\epsilon\}; 0\leq\hat\epsilon\leq\epsilon-\hat\epsilon_g^{\hat k(g)},
    % \end{equation}
    % the rank calibration guarantee to shrink the prediction sets, if for any $e\in E$:
    % \begin{align}\label{al:cond}
    %     \mathbb{P}_{q_{test}}[S(tr(q_{test}, e))\leq\hat s_{\epsilon'}^g, r_{M_\theta}(q_{test}, e)\leq\hat k(g)]\leq\mathbb{P}_{q_{test}}[S(tr(q_{test}, e))\leq\hat s_{\epsilon}^g]
    % \end{align}
    Suppose $\epsilon'(g)$ and $\hat k(g)$ satisfy 
    \begin{equation}
        \hat k(g)\in\{k:\epsilon_g^{\hat k(g)}<\epsilon\}; 0\leq\epsilon'(g)\leq\epsilon-\epsilon_g^{\hat k(g)},
    \end{equation}
    the rank calibration guarantee to shrink the prediction sets, if for a query $q$ and any $e'\in E$:
    \begin{align}\label{al:cond}
        \mathbb{P}_{q}\Big(S(tr(q, e'))\leq\ScoreThreshold, \operatorname{rank}_{M_\theta}(q, e')\leq\hat k(g)\Big)\leq\mathbb{P}_{q}\Big(S(tr(q, e'))\leq\hat s_{\epsilon}(\PartSet)\Big)
    \end{align}
\end{corollary}

\begin{proof}
    We can prove this Corollary based on \citet[Appendix A.2]{shi2024conformal}.Define the following fraction:
    \begin{equation}
        \sigma_g=\frac{ \mathbb{P}_{q}\Big(S(tr(q, e'))\leq\ScoreThreshold, \operatorname{rank}_{M_\theta}(q, e')\leq\RankThreshold\Big)}{\mathbb{P}_{q}(S(tr(q, e'))\leq\hat s_{\epsilon}(\PartSet)}.
    \end{equation}
    By the assumption in Equation (\ref{al:cond}), it follows that $\sigma_g\leq 1$. 
    The expected size of the prediction set for {\CondKGCP} is given by:
    \begin{align}
        \mathbb{E}_{q}\Big[|\PredicitonSetCondKGCP(q)|\Big] &= \mathbb{E}_{q}\bigg[\sum_{e'\in E}\mathbbm{1}\Big[S(tr(q, e'))\leq\ScoreThreshold, \operatorname{rank}_{M_\theta}(q, e')\leq\RankThreshold\Big]\bigg]\\
        &= \sum_{e'\in E}\mathbb{E}_{q}\bigg[\mathbbm{1}\Big[S(tr(q, e'))\leq\ScoreThreshold, \operatorname{rank}_{M_\theta}(q, e')\leq\RankThreshold\Big]\bigg]\\
        &= \sum_{e'\in E}\mathbb{P}_{q}\bigg(S(tr(q, e'))\leq\ScoreThreshold, \operatorname{rank}_{M_\theta}(q, e')\leq\RankThreshold\bigg)
    \end{align}
    By the definition of $\sigma_g$ and the assumption $\sigma_g\leq 1$, we obtain:
    \begin{align}
        \mathbb{E}_{q}\Big[|\PredicitonSetCondKGCP(q)|\Big] &= \sum_{e'\in E}\sigma_g\cdot\mathbb{P}_{q}\Big(S(tr(q, e'))\leq\hat s_{\epsilon}(\PartSet)\Big)\\
        &\leq\sum_{e'\in E}\mathbb{E}_{q}\bigg[\mathbbm{1}\Big[S(tr(q, e'))\leq\hat s_{\epsilon}(\PartSet)\Big]\bigg]\\
        &=\mathbb{E}_{q}\bigg[\sum_{e'\in E}\mathbbm{1}\Big[S(tr(q, e'))\leq\hat s_{\epsilon}(\PartSet)\Big]\bigg]\\
        &=\mathbb{E}_{q}\Big[|\hat C_{\operatorname{MCP^*}}(q)|\Big]
    \end{align}
    Therefore, we conclude that adding rank calibration always reduces the prediction set size, provided that the condition in Equation (\ref{al:cond}) holds.
\end{proof}

\twocolumn

\section{Discussion About Merging Process}
Note the underlying idea of predicate merging procedure in section \ref{sec:merge} is essentially very similar to {\ClusterCP} proposed by recent work \cite{ding2024class}, where subgroups are clustered based on similarity of the nonconformity score distribution. But in our case, we have extremely imbalanced size of calibration data for each predicate. For many predicates, the number of calibration triples is too small to capture the characteristics of score distribution with quantile vectors proposed in \citet{ding2024class}, thus resulting unreliable clustering and limited improvement in terms of conditional coverage probability. We can see in Table \ref{tab:main_filter} that {\ClusterCP} fails to provide meaningful fine-grained predicate clustering, reflected by very similar performance as {\KGCP} (baseline method that is not designed to achieve conditional coverage guarantees). 

\section{Verification of the Condition in Equation (\ref{eq:cond})}\label{app:verify}
As we prove in Appendix \ref{app:proof}, {\CondKGCP} always has smaller expected prediction set size compared to {\CondKGCP} without rank calibration, i.e., {\MCP} at set part level under the condition in Equation (\ref{eq:cond}). In this section, we empirically verify the condition in Equation (\ref{eq:cond}) across all model-dataset combinations.

We use two metrics to verify the condition.
Condition Satisfaction Rate (CSR) quantifies how often the condition $\sigma_g\leq 1$ holds for all $g\in P$:
\begin{equation}
    CSR:=\sum_{g\in P}\mathbbm{1}[\sigma_g\leq 1];
\end{equation}
And $\bar\sigma$ computes the average value of $\sigma_g$ for all $g\in P$, denoted as $\bar\sigma$:
\begin{equation}
    \bar\sigma:=\frac{1}{|P|}\sum_{g\in P}\sigma_g.
\end{equation}

The results of these two metrics for all model-dataset combinations are reported in Table \ref{tab:condition}. The results show that the condition $\sigma_g\leq 1$ holds for nearly all $g\in P$ under different model-dataset combinations. This provides empirical evidence that the additional rank calibration schema reduces the size of prediction sets.
Moreover, we observe that smaller values of $\bar\sigma$ are typically associated with smaller {\AvgSize} in in Table \ref{tab:main_filter}.

\begin{table}[t!]
\resizebox{0.45\textwidth}{!}{%
\begin{tabular}{ccc|cc}
\toprule
 & \multicolumn{2}{c}{WN18} & \multicolumn{2}{c}{FB15k} \\
Model & CSR (\%) & $\bar\sigma$ & CSR (\%) & $\bar\sigma$ \\\midrule
TransE & 91.7 & 0.823 & 98.7 & 0.633 \\
RotatE & 100 & 0.422 & 97.6 & 0.899 \\
RESCAL & 100 & 0.554 & 99.3 & 0.712 \\
DistMult & 91.7 & 0.783 & 99.4 & 0.512 \\
ComplEx & 100 & 0.401 & 96.4 & 0.703 \\
ConvE & 100 & 0.555 & 93.1 & 0.951\\
\bottomrule
\end{tabular}%
}
\caption{Verification of Condition in Equation (\ref{eq:cond}).}
\label{tab:condition}
\end{table}

\section{SOFTMAX Nonconformity Score}\label{app:default_score}
By default we use {\Softmax} nonconformity score defined in \citet{zhu2024conformalized}:
\begin{equation*}
    S(tr(q,e)) = 1-\hat M_{\theta}(tr(q,e)),
\end{equation*}
where
\begin{equation*}
    \hat M_{\theta}(tr(q,e))=\frac{\exp(M_{\theta}(tr(q,e)))}{\sum_{e'\in E}\exp{(M_{\theta}(q,e'))}}.
\end{equation*}

\section{APS and RAPS Nonconformity Score}\label{app:advanced_scores}
Adaptive Predication Sets ({\APS}) \cite{romano2020classification} improves the conditional coverage probability by modifying the nonconformity measure. Specifically, given a query $q$, for all $e\in E$, we first normalize the plausibility score using softmax function:
\begin{equation*}
    \hat M_{\theta}(tr(q,e))=\frac{\exp(M_{\theta}(tr(q,e)))}{\sum_{e'\in E}\exp{(M_{\theta}(q,e'))}}.
\end{equation*}
Then we sort the normalized scores such that $1\geq \hat M_{(1)}\geq\dots\geq\hat M_{(|E|)}$, where $M_{(k)}$ denotes the $k$-th largest plausibility score. Recall that $\operatorname{rank}_{M_\theta}(q,e)$ denotes the rank of $e$ given query $q$. The nonconformity score of {\APS} is then defined as

\begin{align*}
    &S(tr(q,e)) = \\
    &\sum_{i=1}^{\operatorname{rank}_{M_\theta}(q,e)-1}\hat M_{(i)} + U\cdot\hat M_{(\operatorname{rank}_{M_\theta}(q,e))},
\end{align*}
where $U\in[0,1]$ is a uniform random variable.

The regularized version - {\RAPS} additionally includes a rank-based regularization term to the nonconformity score.
\begin{align*}
    &S(tr(q,e)) = \\
    &\sum_{i=1}^{\operatorname{rank}_{M_\theta}(q,e)-1}\hat M_{(i)} + U\cdot\hat M_{(\operatorname{rank}_{M_\theta}(q,e))}\\
    &+\lambda\cdot\max\{\operatorname{rank}_{M_\theta}(q,e)-k_{reg}, 0\},
\end{align*}
where $\lambda$ and $k_{reg}$ are two hyper-parameters.

\section{Detailed Experimental Settings}
Note the experimental settings closely follow the approach outlined by \citet{zhu2024conformalized}. For completeness and to ensure the paper is self-contained, we recall the details in this section.

\subsection{Information About KGE Models and Benchmark Datasets}
Table \ref{tab:dataset} outlines key statistics for the benchmark datasets, while Table \ref{tab:scoring_functions} presents the scoring functions utilized by various KGE methods.

\begin{table}[t!]
\resizebox{.48\textwidth}{!}{%
\begin{tabular}{@{}llllll@{}}
\toprule
          & \#Entity & \#Relation & \#Training & \#Validation & \#Test \\ \midrule
WN18      & 40,943   & 18       & 141,442    & 5,000        & 5,000  \\
WN18RR    & 40,943   & 11       & 86,835     & 3,034        & 3,134  \\
FB15k     & 14,951   & 1,345    & 483,142    & 50,000       & 59,071 \\
FB15k-237 & 14,541   & 237      & 272,115    & 17,535       & 20,466 \\ \bottomrule
\end{tabular}%
}
\caption{Statistics of benchmark datasets for link prediction task.}
\label{tab:dataset}
\end{table}

\begin{table}[t!]
\centering
\resizebox{.48\textwidth}{!}{%
\begin{tabular}{lc}
\toprule
         & Scoring Function $s(<h,r,t>)$\\ \midrule
TransE \cite{bordes2013translating}   & $-||\mathbf{h}+\mathbf{r}-\mathbf{t}||_{1/2}$\\
RotatE \cite{sun2019rotate}  & $-||\mathbf{h}\circ\mathbf{r}-\mathbf{t}||_{p}$\\
RESCAL \cite{nickel2011rescal}   & $\mathbf{h}^T\mathbf{M}_r\mathbf{t}$\\
DistMult \cite{yang2015distmult} & $\mathbf{h}^Tdiag(\mathbf{r})\mathbf{t}$\\
ComplEx \cite{trouillon2016complex}  & $Re(\mathbf{h}^Tdiag(\mathbf{r})\overline{\mathbf{t}})$\\
ConvE \cite{dettmers2018convolutional}   & $f(vec(f([\overline{\mathbf{h}};\overline{\mathbf{r}}]*\omega))\mathbf{W})\mathbf{t}$\\ \bottomrule
\end{tabular}%
}
\caption{The scoring function $s(<\boldsymbol{h,r,t}>)$ of KGE models used in this paper, where $\boldsymbol{h,r,t}$ denote the embeddings of $h,r,t$, $\circ$ denotes Hadamard product. $\overline{\cdot}$ refers to conjugate for complex vectors in ComplEx, and 2D reshaping for real vectors in ConvE. $*$ is operator for 2D convolution. $\omega$ is the filters and $W$ is the parameters for 2D convolutional layer.}
\label{tab:scoring_functions}
\end{table}

\subsection{Privacy Concerns in FB15k and FB15k-237}
Both FB15k and FB15k-237 datasets include data about individuals, predominantly well-known public figures such as politicians, celebrities, and historical icons. Since this information is widely accessible through public platforms and online sources, its inclusion in Freebase poses minimal privacy risks compared to datasets containing sensitive or private personal details.

\subsection{Details of Pre-training KGE Models}
For pre-training the the KGE models, we follow \cite{zhu2024conformalized}.

The LibKGE framework \cite{libkge} was used for training the KGE models, following a hyperparameter optimization approach inspired by \cite{ruffinelli2019you}. The experiments were executed on a Linux system equipped with a 40GB NVIDIA A100 SXM4 GPU.

Initially, we applied a quasi-random hyperparameter search using Sobol sequences to ensure an even distribution of configurations, avoiding clustering \cite{bergstra2012random}. For each combination of dataset, model, training type, and loss function, 30 configurations were generated. This was followed by a Bayesian optimization phase, incorporating 30 additional trials to refine the hyperparameters based on prior results. The Ax framework (\url{https://ax.dev/}) facilitated this process.

The search spanned a comprehensive hyperparameter space, encompassing loss functions (pairwise margin ranking with hinge loss, binary cross-entropy, cross-entropy), regularization methods (none/L1/L2/L3, dropout), optimizers (Adam, Adagrad), and common initialization techniques used in the KGE domain. Embedding sizes of 128, 256, and 512 were considered. For further details, refer to \cite[Table 5]{ruffinelli2019you}.

Configuration files for baseline and competing models, along with models used in aggregation, are available in the "configs" folder of the submitted software directory. These files (*.yaml) document the hyperparameter settings applied in this study.

\begin{table*}[t!]
\centering
\resizebox{\textwidth}{!}{%
\begin{tabular}{cccc|cccc}
\toprule
\multicolumn{4}{c}{WN18} & \multicolumn{4}{c}{FB15k} \\
Method & Training & Calibration & Set Construction & Method & Training & Calibration & Set Construction \\\midrule
{\KGCP} & 1h & 6.68s & 0.8ms/query & {\KGCP} & 2h & 32.70s & 0.4ms/query \\
{\MCP} & 1h & 7.32s & 0.8ms/query & {\MCP} & 2h & 32.70s & 0.4ms/query \\
{\ClusterCP} & 1h & 8.30s & 0.8ms/query & {\ClusterCP} & 2h & 33.55s & 0.4ms/query \\
{\APS} & 1h & 7.28s & 1ms/query & {\APS} & 2h & 32.56s & 0.4ms/query \\
{\RAPS} & 1h & 8.72s & 1ms/query & {\RAPS} & 2h & 39.23s & 0.5ms/query \\
{\CondKGCP} & 1h & 8.92s & 1ms/query & {\CondKGCP} & 2h & 38.33s & 0.5ms/query \\
\bottomrule
\end{tabular}%
}
\caption{Empirical runtime for different uncertainty quantification methods on WN18 and FB15k.}
\label{tab:runtime}
\end{table*}

\subsection{Optimal Hyperparameter Settings}
The optimal hyperparameter configurations for each model-dataset combination are summarized in Table \ref{tab:condkgcp_hyper}.

\begin{table}[t!]
\centering
\resizebox{0.38\textwidth}{!}{%
\begin{tabular}{ccc|cc}
\toprule
 & \multicolumn{2}{c}{WN18} & \multicolumn{2}{c}{FB15k} \\
Model & gamma & cut & gamma & cut \\\midrule
TransE & 0.01 & 50 & 0.01 & 20 \\
RotatE & 0.01 & 50 & 0.01 & 100 \\
RESCAL & 0.01 & 50 & 0.01 & 50 \\
DistMult & 0.01 & 50 & 0.01 & 20 \\
ComplEx & 0.1 & 50 & 0.01 & 20 \\
ConvE & 0.01 & 50 & 0.01 & 50\\
\bottomrule
\end{tabular}%
}
\caption{Optimal hyperparameter configurations for {\CondKGCP} across various model-dataset combinations.}
\label{tab:condkgcp_hyper}
\end{table}

\section{Complete Experimental Results}
\begin{figure*}[t!]
    \centering
    \includegraphics[width=\textwidth]{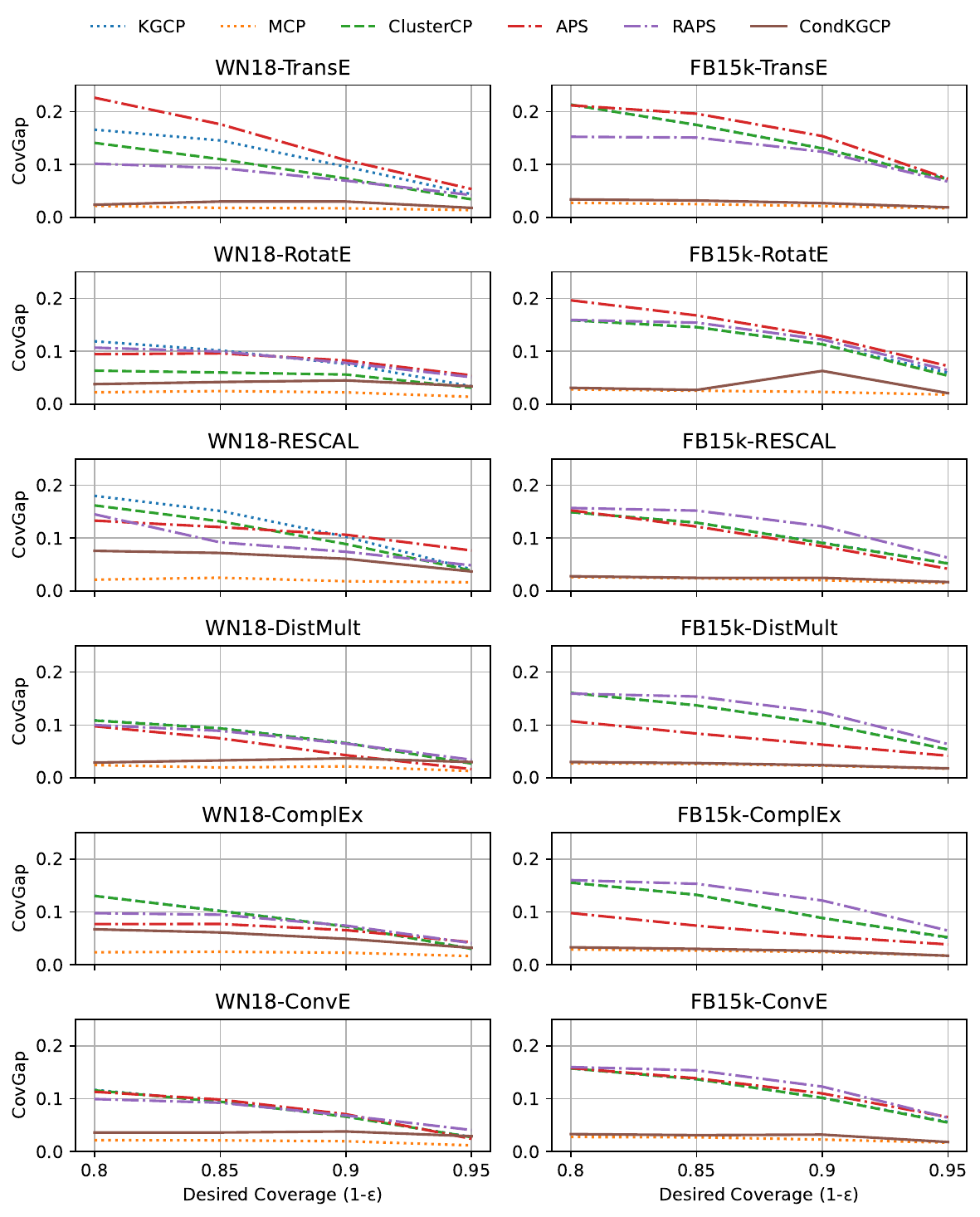}
    \caption{Complete results of comparison of methods' {\CovGap} across varying target coverage levels.}
    \label{fig:alpha_covgap}
\end{figure*}

\begin{figure*}[t!]
    \centering
    \includegraphics[width=\textwidth]{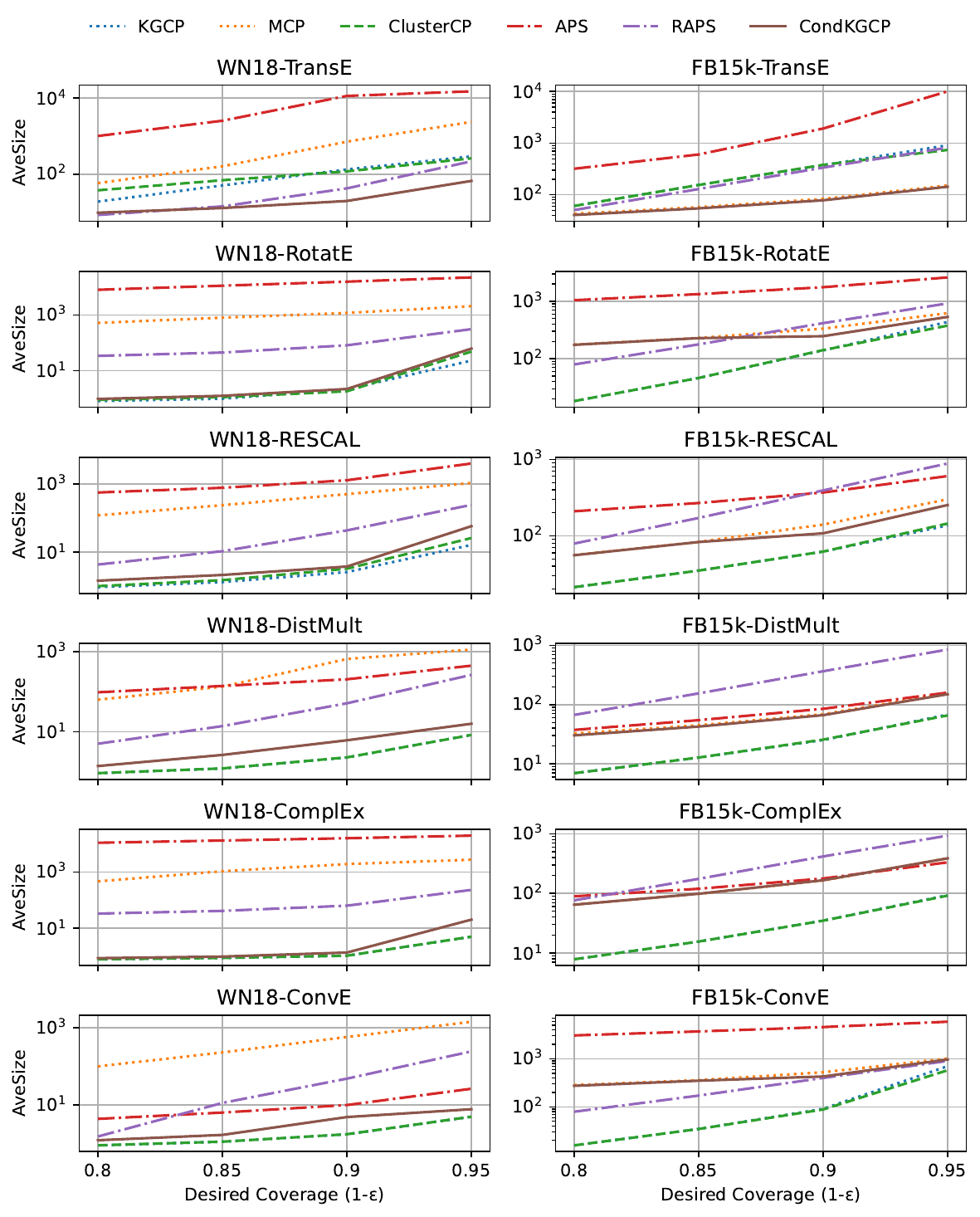}
    \caption{Complete results of comparison of methods' {\AvgSize} across varying target coverage levels.}
    \label{fig:alpha_size}
\end{figure*}

In Figure \ref{fig:alpha_covgap} and \ref{fig:alpha_size}, we show the complete results of comparison of methods' {\CovGap} and {\AvgSize} across varying target coverage levels ranging from $[0.8, 0.85, 0.9, 0.95]$.

\begin{figure*}[t!]
    \centering
    \includegraphics[width=\textwidth]{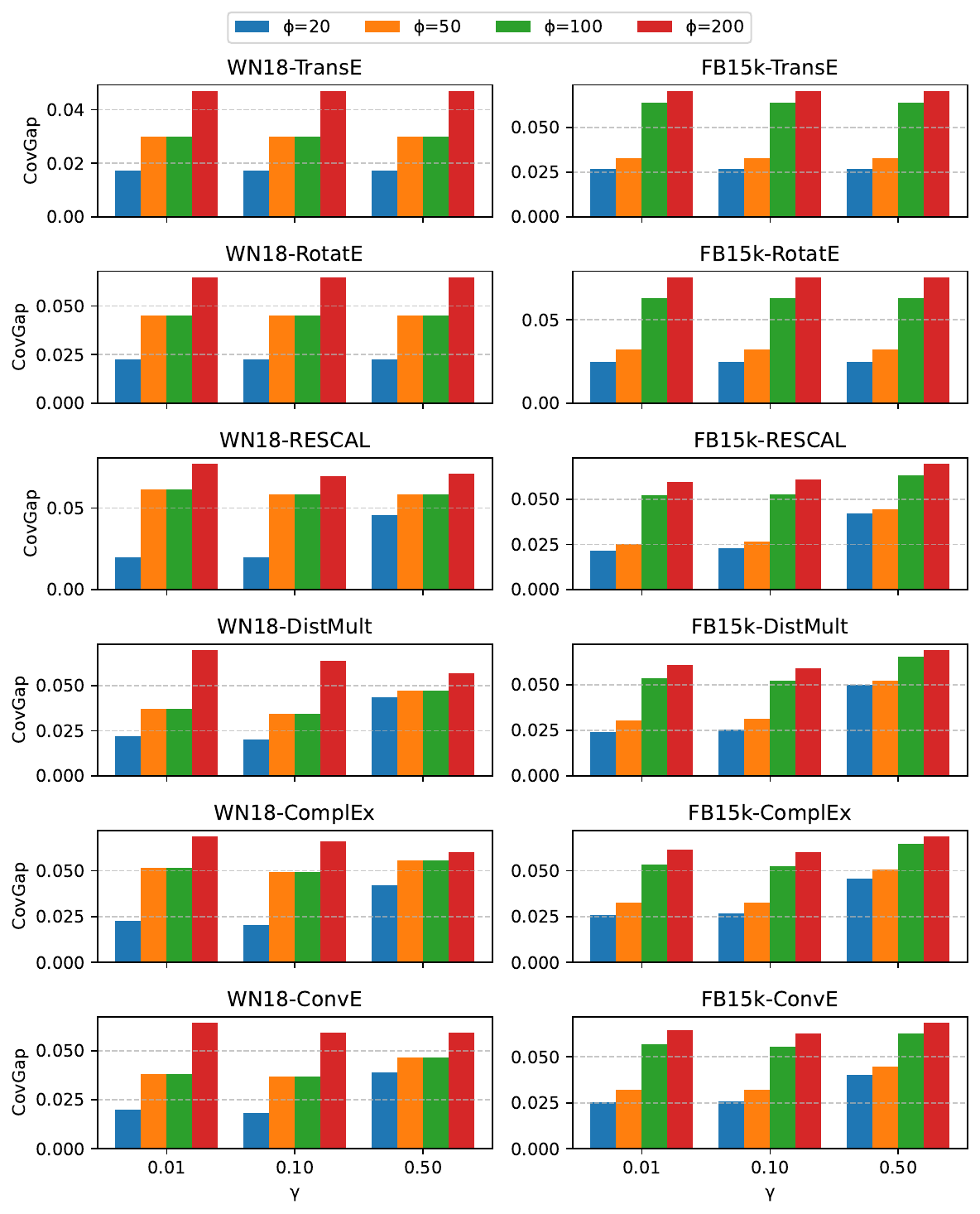}
    \caption{Influence of hyperparameters $\phi$ and $\gamma$ on {\CovGap}.}
    \label{fig:hyper_covgap}
\end{figure*}

\begin{figure*}[t!]
    \centering
    \includegraphics[width=\textwidth]{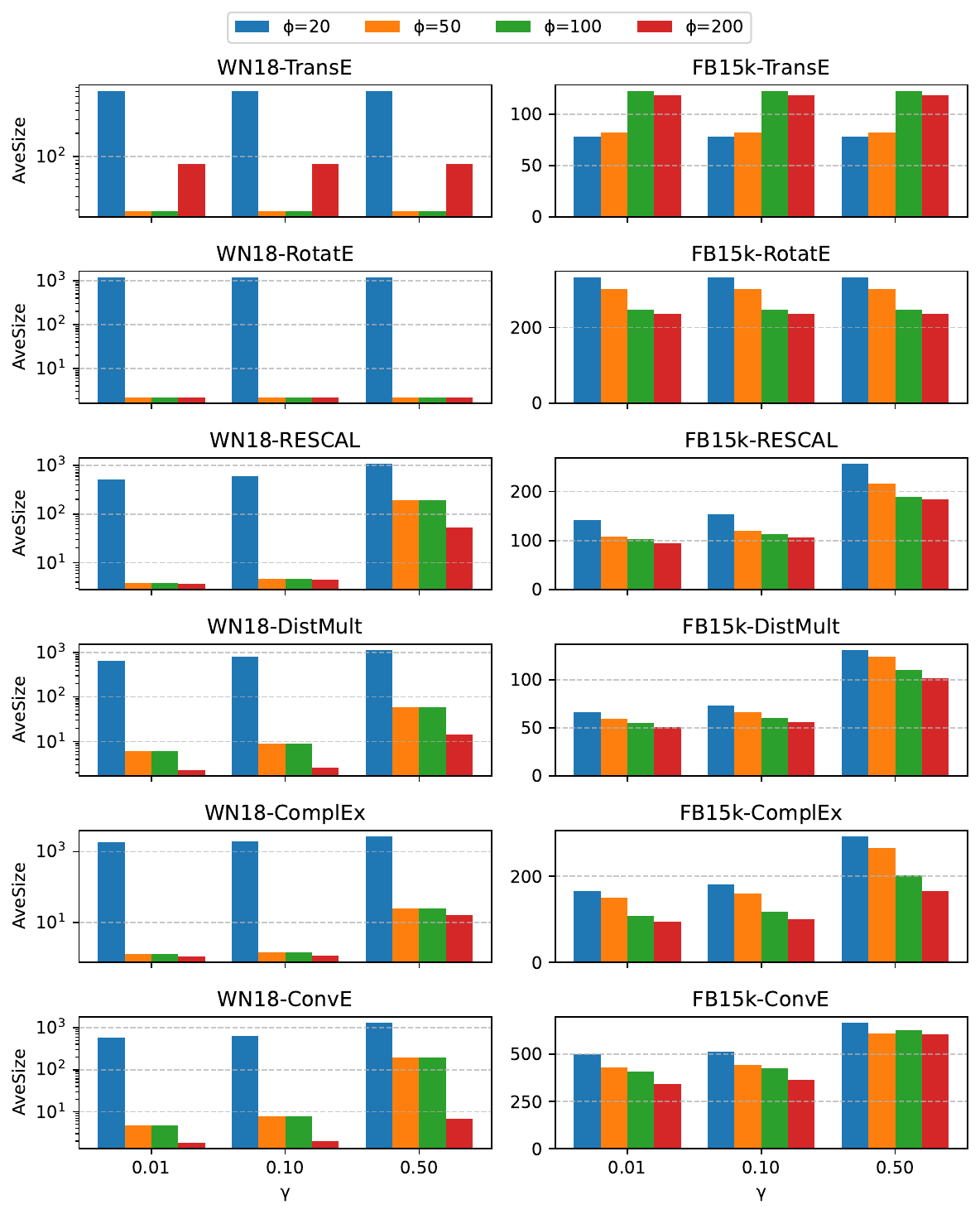}
    \caption{Influence of hyperparameters $\phi$ and $\gamma$ on {\AvgSize}.}
    \label{fig:hyper_size}
\end{figure*}

In Figure \ref{fig:hyper_covgap} and \ref{fig:hyper_size}, we investigate the influence of hyperparameters $\phi$ and $\gamma$ on {\CovGap} and {\AvgSize}, respectively.

\section{Complexity Analysis}

The total computational cost of our method, {\CondKGCP}, comprises three main components:
\begin{itemize}
    \item \textbf{KGE Model Training}: This is shared across all methods and is the most computationally intensive component.
    \item \textbf{Calibration Step}: This includes any method-specific operations, such as clustering in {\ClusterCP} or dual calibration in {\CondKGCP}. These are one-time offline steps and incur negligible overhead relative to model training.
    \item \textbf{Test-Time Set Construction}: This is the only component executed per test query and constitutes the primary source of runtime differences between methods.
\end{itemize}

We now present a theoretical complexity analysis focused on the per-query cost:

\begin{itemize}
    \item Baseline methods (e.g., {\KGCP}, {\MCP}): 
    Each query requires computing scores for all candidate entities, with time complexity $\mathcal{O}(|E|\cdot d)$, where $|E|$ is the number of entities and $d$ is the embedding dimension. An additional $\mathcal{O}(|E|)$ is needed for score thresholding to form the prediction set.
    \item {\CondKGCP}: In addition to score computation ($\mathcal{O}(|E|\cdot d)$), our method includes rank-based thresholding from dual calibration, adding a sorting step $\mathcal{O}(|E|\log|E|)$, linear pass $\mathcal{O}(|E|)$, and selection of the top-$K$ entities ($\mathcal{O}(K)$). Thus, the total per-query complexity is: $\mathcal{O}(|E|\cdot d) + \mathcal{O}(|E|\log|E| + |E| + K)$.
\end{itemize}

While {\CondKGCP} introduces additional steps beyond baselines, the dominant term remains $\mathcal{O}(|E|\cdot d)$ in practice. For instance, with $|E| = 10^6$ and $d = 512$, computing scores dominates (512 operations per entity), whereas the extra overhead from sorting and filtering (approximately 21 operations per entity) is comparatively negligible.

Empirical runtimes, reported in Table~\ref{tab:runtime}, confirm that the additional complexity of {\CondKGCP} does not translate into significant runtime overhead on a NVIDIA A100 GPU.

\section{Results for More Confidence Levels}

\begin{table}[t]
\resizebox{\linewidth}{!}{%
\centering
\begin{tabular}{cccc}
\toprule
 & \multicolumn{1}{l}{CovGap (0.95)} & \multicolumn{1}{l}{AveSize (0.95)} & Ratio (0.95) \\\midrule
{\KGCP} & 0.040 & 16.30 & -- \\
{\MCP} & 0.013 & 1128.46 & \multicolumn{1}{r}{41191.11} \\
{\ClusterCP} & 0.040 & 19.61 & -- \\
{\APS} & 0.079 & 4036.71 & -- \\
{\RAPS} & 0.051 & 242.36 & -- \\
{\CondKGCP} & 0.029 & 133.37 & \multicolumn{1}{r}{10642.73} \\\midrule
 & \multicolumn{1}{l}{CovGap (0.98)} & \multicolumn{1}{l}{AveSize (0.98)} & Ratio (0.98) \\\midrule
{\KGCP} & 0.020 & 2945.21 & -- \\
{\MCP} & 0.009 & 5027.66 & \multicolumn{1}{r}{189313.64} \\
{\ClusterCP} & 0.022 & 2933.73 & -- \\
{\APS} & 0.039 & 12094.28 & -- \\
{\RAPS} & 0.018 & 3985.23 & \multicolumn{1}{r}{520010.00} \\
{\CondKGCP} & 0.017 & 3060.69 & \multicolumn{1}{r}{38493.33} \\\midrule
 & \multicolumn{1}{l}{CovGap (0.99)} & \multicolumn{1}{l}{AveSize (0.99)} & Ratio (0.99) \\\midrule
{\KGCP} & 0.016 & 7403.45 & -- \\
{\MCP} & 0.006 & 9494.30 & \multicolumn{1}{r}{209085.00} \\
{\ClusterCP} & 0.018 & 8921.37 & -- \\
{\APS} & 0.016 & 19486.37 & -- \\
{\RAPS} & 0.009 & 17098.86 & \multicolumn{1}{r}{1385058.57} \\
{\CondKGCP} & 0.013 & 7614.11 & \multicolumn{1}{r}{70220.00}\\
\bottomrule
\end{tabular}%
}
\caption{Performance results at high confidence levels ($1-\epsilon > 0.95$).}
\label{tab:highconfidence}
\end{table}

In Table~\ref{tab:highconfidence}, we extend the analysis from Figure~\ref{fig:alpha_inves} to higher confidence levels (i.e., $\alpha > 0.95$) using the same experimental settings. We observe that {\CondKGCP} consistently achieves superior conditional coverage while maintaining a more favorable trade-off between coverage and prediction set size, even at these stringent confidence levels.

Notably, all methods exhibit a sharp increase in prediction set size beyond $\alpha = 0.95$. We attribute this to the inherent limitations of the base KGE models. For instance, on WN18, most models attain Hits@10 close to 0.95, meaning that approximately 95\% of correct answers are ranked within the top 10. Pushing coverage beyond this threshold necessitates including lower-ranked (and often noisy) entities, which substantially enlarges the prediction sets.

This observation underscores an important practical consideration: in high-stakes applications where coverage above 95\% is required, it is essential to pair uncertainty quantification with a highly accurate base model (e.g., achieving Hits@$K$ > 0.95). Otherwise, the resulting prediction sets may become impractically large.

\section{AI Assistants In Writing}
We use ChatGPT \cite{openai2024chatgpt} to enhance our writing skills, abstaining from its use in research and coding endeavors.

\begin{table*}[t!]
\centering
\resizebox{\textwidth}{!}{%
\begin{tabular}{cccrr||cccrr}
\toprule[2pt]
\multicolumn{5}{c}{WN18} & \multicolumn{5}{c}{FB15k} \\
Model & Score & Method & CovGap $\downarrow$ & AveSize $\downarrow$ & Model & Score & Method & CovGap $\downarrow$ & AveSize $\downarrow$ \\\midrule
\multirow{6}{*}{TransE} & \multirow{3}{*}{{\APS}} & {\MCP} & 0.015 & 1320.77 & \multirow{6}{*}{TransE} & \multirow{3}{*}{{\APS}} & {\MCP} & 0.022 & 653.62 \\
 &  & {\ClusterCP} & 0.079 & 9081.71 &  &  & {\ClusterCP} & 0.116 & 4925.66 \\
 &  & {\CondKGCP} & 0.016 & 1090.23 &  &  & {\CondKGCP} & 0.028 & 158.17 \\\cmidrule(lr){2-5}\cmidrule(lr){7-10}
 & \multirow{3}{*}{{\RAPS}} & {\MCP} & 0.024 & 1172.74 &  & \multirow{3}{*}{{\RAPS}} & {\MCP} & 0.021 & 566.15 \\
 &  & {\ClusterCP} & 0.066 & 1138.93 &  &  & {\ClusterCP} & 0.122 & 320.57 \\
 &  & {\CondKGCP} & 0.051 & 50.29 &  &  & {\CondKGCP} & 0.023 & 258.90 \\\midrule
 
\multirow{6}{*}{RotatE} & \multirow{3}{*}{{\APS}} & {\MCP} & 0.014 & 15922.23 & \multirow{6}{*}{RotatE} & \multirow{3}{*}{{\APS}} & {\MCP} & 0.022 & 1695.87 \\
 &  & {\ClusterCP} & 0.077 & 15398.99 &  &  & {\ClusterCP} & 0.122 & 1626.64 \\
 &  & {\CondKGCP} & 0.014 & 15920.71 &  &  & {\CondKGCP} & 0.023 & 1389.67 \\\cmidrule(lr){2-5}\cmidrule(lr){7-10}
 & \multirow{3}{*}{{\RAPS}} & {\MCP} & 0.022 & 2084.58 &  & \multirow{3}{*}{{\RAPS}} & {\MCP} & 0.022 & 523.75 \\
 &  & {\ClusterCP} & 0.063 & 2051.17 &  &  & {\ClusterCP} & 0.121 & 398.75 \\
 &  & {\CondKGCP} & 0.039 & 141.42 &  &  & {\CondKGCP} & 0.023 & 322.63 \\\midrule
 
\multirow{6}{*}{RESCAL} & \multirow{3}{*}{{\APS}} & {\MCP} & 0.020 & 3236.86 & \multirow{6}{*}{RESCAL} & \multirow{3}{*}{{\APS}} & {\MCP} & 0.021 & 867.08 \\
 &  & {\ClusterCP} & 0.095 & 1932.13 &  &  & {\ClusterCP} & 0.072 & 360.10 \\
 &  & {\CondKGCP} & 0.020 & 3334.02 &  &  & {\CondKGCP} & 0.024 & 552.58 \\\cmidrule(lr){2-5}\cmidrule(lr){7-10}
 & \multirow{3}{*}{{\RAPS}} & {\MCP} & 0.020 & 1037.90 &  & \multirow{3}{*}{{\RAPS}} & {\MCP} & 0.021 & 590.23 \\
 &  & {\ClusterCP} & 0.060 & 1001.70 &  &  & {\ClusterCP} & 0.121 & 374.67 \\
 &  & {\CondKGCP} & 0.035 & 94.29 &  &  & {\CondKGCP} & 0.025 & 287.15 \\\midrule
 
\multirow{6}{*}{DistMult} & \multirow{3}{*}{{\APS}} & {\MCP} & 0.018 & 815.22 & \multirow{6}{*}{DistMult} & \multirow{3}{*}{{\APS}} & {\MCP} & 0.024 & 723.68 \\
 &  & {\ClusterCP} & 0.043 & 205.65 &  &  & {\ClusterCP} & 0.063 & 85.04 \\
 &  & {\CondKGCP} & 0.019 & 249.74 &  &  & {\CondKGCP} & 0.024 & 124.64 \\\cmidrule(lr){2-5}\cmidrule(lr){7-10}
 & \multirow{3}{*}{{\RAPS}} & {\MCP} & 0.022 & 1003.06 &  & \multirow{3}{*}{{\RAPS}} & {\MCP} & 0.022 & 576.24 \\
 &  & {\ClusterCP} & 0.059 & 969.60 &  &  & {\ClusterCP} & 0.123 & 356.66 \\
 &  & {\CondKGCP} & 0.024 & 140.75 &  &  & {\CondKGCP} & 0.023 & 273.34 \\\midrule
 
\multirow{6}{*}{ComplEx} & \multirow{3}{*}{{\APS}} & {\MCP} & 0.017 & 16707.74 & \multirow{6}{*}{ComplEx} & \multirow{3}{*}{{\APS}} & {\MCP} & 0.024 & 694.64 \\
 &  & {\ClusterCP} & 0.065 & 15738.09  &  &  & {\ClusterCP} & 0.054 & 176.78 \\
 &  & {\CondKGCP} & 0.017 & 16728.70 &  &  & {\CondKGCP} & 0.025 & 282.54  \\\cmidrule(lr){2-5}\cmidrule(lr){7-10}
 & \multirow{3}{*}{{\RAPS}} & {\MCP} & 0.023 & 2614.74 &  & \multirow{3}{*}{{\RAPS}} & {\MCP} & 0.023 & 530.80 \\
 &  & {\ClusterCP} & 0.064 & 2598.85 &  &  & {\ClusterCP} & 0.120 & 401.65 \\
 &  & {\CondKGCP} & 0.041 & 83.65 &  &  & {\CondKGCP} & 0.026 & 331.38 \\\midrule
 
\multirow{6}{*}{ConvE} & \multirow{3}{*}{{\APS}} & {\MCP} & 0.015 & 589.50 & \multirow{6}{*}{ConvE} & \multirow{3}{*}{{\APS}} & {\MCP} & 0.022 & 3951.64  \\
 &  & {\ClusterCP} & 0.070 & 9.91  &  &  & {\ClusterCP} & 0.109 & 4376.19  \\
 &  & {\CondKGCP} & 0.052 & 11.89  &  &  & {\CondKGCP} & 0.025 & 3019.98  \\\cmidrule(lr){2-5}\cmidrule(lr){7-10}
 & \multirow{3}{*}{{\RAPS}} & {\MCP} & 0.023 & 1013.09 &  & \multirow{3}{*}{{\RAPS}} & {\MCP} & 0.022 & 613.67  \\
 &  & {\ClusterCP} & 0.060 & 981.74  &  &  & {\ClusterCP} & 0.121 & 386.85  \\
 &  & {\CondKGCP} & 0.033 & 116.70  &  &  & {\CondKGCP} & 0.025 & 304.28 \\
 \bottomrule[2pt]
\end{tabular}%
}
\caption{Overall performance comparison of {\MCP}, {\ClusterCP}, and {\CondKGCP} using different nonconformity measures: {\APS} and {\RAPS}. Note that methods modifying nonconformity scores theoretically can be combined with subgroup-based methods, however since {\APS} and {\RAPS} are not suitable for KGE methods we do not see much improvement by combining these methods, nevertheless, {\CondKGCP} outperforms both {\MCP} and {\ClusterCP} if we combine with {\APS} or {\RAPS} across all evaluation metrics.}
\label{tab:score_results}
\end{table*}

\end{document}